\ifdefined\pdfinfoomitdate     \pdfinfoomitdate=1      \fi
\ifdefined\pdftrailerid        \pdftrailerid{}         \fi
\ifdefined\pdfsuppressptexinfo \pdfsuppressptexinfo=-1 \fi
\documentclass[prx,superscriptaddress,aps,amsmath,amssymb,floatfix,reprint,raggedbottom,longbibliography]{revtex4-2}
\usepackage{graphicx}
\usepackage{enumitem}
\usepackage{amssymb}
\usepackage{silence}
\usepackage{float}

\usepackage[ruled,lined,linesnumbered]{algorithm2e}
\usepackage{balance}
\usepackage{silence}
\usepackage{hyperref}
\hypersetup{
    colorlinks=true,       
    linkcolor=red,          
  citecolor=magenta,        
    filecolor=magenta,      
    urlcolor=red,           
    runcolor=cyan
}
\usepackage{amsmath}
\DeclareMathOperator{\arctanh}{arctanh}
\usepackage{times}

\usepackage{dcolumn}
\usepackage{booktabs}

\usepackage{xcolor}
\usepackage{lipsum}
\usepackage{soul}
\usepackage{braket}
\usepackage{amsfonts}
\usepackage{multirow}
\usepackage[utf8]{inputenc}
\DeclareUnicodeCharacter{2009}{\,}

\makeatletter
\newcommand*{\addFileDependency}[1]{
  \typeout{(#1)}
  \@addtofilelist{#1}
  \IfFileExists{#1}{}{\typeout{No file #1.}}
}
\makeatother

\makeatletter 
\renewcommand{\fnum@figure}{\textbf{Fig.~\thefigure}}
\makeatother

\makeatletter
\def\bbordermatrix#1{\begingroup \m@th
  \@tempdima 4.75\p@
  \setbox\z@\vbox{%
    \def\cr{\crcr\noalign{\kern2\p@\global\let\cr\endline}}%
    \ialign{$##$\hfil\kern2\p@\kern\@tempdima&\thinspace\hfil$##$\hfil
      &&\quad\hfil$##$\hfil\crcr
      \omit\strut\hfil\crcr\noalign{\kern-\baselineskip}%
      #1\crcr\omit\strut\cr}}%
  \setbox\tw@\vbox{\unvcopy\z@\global\setbox\@ne\lastbox}%
  \setbox\tw@\hbox{\unhbox\@ne\unskip\global\setbox\@ne\lastbox}%
  \setbox\tw@\hbox{$\kern\wd\@ne\kern-\@tempdima\left[\kern-\wd\@ne
    \global\setbox\@ne\vbox{\box\@ne\kern2\p@}%
    \vcenter{\kern-\ht\@ne\unvbox\z@\kern-\baselineskip}\,\right]$}%
  \null\;\vbox{\kern\ht\@ne\box\tw@}\endgroup}
\makeatother

\emergencystretch=\maxdimen
\usepackage[compact]{titlesec}
\titlespacing{\section}{0pt}{*3}{*2}
\titlespacing{\subsection}{0pt}{*2}{*2}
\titlespacing{\subsubsection}{0pt}{*2}{*2}

\titleformat{\section}{\filcenter\normalfont\small \bfseries}{\thesection.}{1em}{\MakeUppercase}   

\newcommand{\beginsupplement}{%
        \setcounter{section}{0}
        \setcounter{table}{0}
        \renewcommand{\thetable}{S\arabic{table}}%
        \setcounter{figure}{0}
        \renewcommand{\thefigure}{S\arabic{figure}}%
        \setcounter{equation}{0}
        \renewcommand{\theequation}{S.\arabic{equation}}%
     }

\begin{document}
\title{From Independent to Correlated Diffusion: \texorpdfstring{\\}{ } Generalized Generative Modeling with Probabilistic Computers}

\author{Nihal Sanjay Singh}
\email{nihalsingh@ece.ucsb.edu}
\affiliation{Department of Electrical and Computer Engineering, University of California, Santa Barbara, Santa Barbara, CA, 93106, USA}

\author{Mazdak Mohseni-Rajaee}
\affiliation{Department of Electrical and Computer Engineering, University of California, Santa Barbara, Santa Barbara, CA, 93106, USA}

\author{Shaila Niazi}
\affiliation{Department of Electrical and Computer Engineering, University of California, Santa Barbara, Santa Barbara, CA, 93106, USA}

\author{Kerem Y. Camsari}\email{camsari@ece.ucsb.edu}
\affiliation{Department of Electrical and Computer Engineering, University of California, Santa Barbara, Santa Barbara, CA, 93106, USA}

\begin{abstract}
Diffusion models have emerged as a powerful framework for generative tasks in deep learning. They decompose generative modeling into two computational primitives: deterministic neural-network evaluation and stochastic sampling. Current implementations usually place most computation in the neural network, but diffusion as a framework allows a broader range of choices for the stochastic transition kernel. Here, we generalize the stochastic sampling component by replacing independent noise injection with Markov chain Monte Carlo (MCMC) dynamics that incorporate known interaction structure. Standard independent diffusion is recovered as a special case when couplings are set to zero. By explicitly incorporating Ising couplings into the diffusion dynamics, the noising and denoising processes exploit spatial correlations representative of the target system. The resulting framework maps naturally onto probabilistic computers (p-computers) built from probabilistic bits (p-bits), which provide orders-of-magnitude advantages in sampling throughput and energy efficiency over GPUs. We demonstrate the approach on equilibrium states of the 2D ferromagnetic Ising model and the 3D Edwards-Anderson spin glass, showing that correlated diffusion produces samples in closer agreement with MCMC reference distributions than independent diffusion. More broadly, the framework shows that p-computers can enable new classes of diffusion algorithms that exploit structured probabilistic sampling for generative modeling. \end{abstract}
\maketitle

\section{Introduction}
\label{sec:Intro}

Diffusion models are a class of generative models inspired by non-equilibrium thermodynamics~\cite{dickstein2015diffusion}.
They operate through a forward \textit{stochastic noising} process that progressively corrupts data, a \textit{training} phase in which a neural network learns to reverse this corruption, and \textit{inference}, in which new samples are generated by iteratively applying learned, stochastic denoising transitions~\cite{ho2020ddpm,austin2021d3pm}.
The probabilistic nature of diffusion models means they are inherently composed of two computational components: deterministic neural-network evaluation (a conditional estimator) and stochastic sampling (a transition kernel that realizes the next state).

A central but often overlooked aspect of diffusion models is that their algorithmic structure imposes no fundamental constraint on how computational effort is divided between these two components.
In current practice, the widespread availability of GPUs has pushed most of the cost into the neural-network evaluation, which is well matched to dense linear algebra, while the stochastic sampling is typically kept lightweight to fit what this hardware makes easiest to execute~\cite{sara2020hardlott}.
This is a hardware-contingent design choice rather than a property of the diffusion framework, and it leaves underexplored a complementary regime in which the sampling step is fast enough to justify richer, more informative transition kernels.

Probabilistic computers (p-computers) built from probabilistic bits (p-bits)~\cite{camsari2019pbits, chowdhury2023fullstack} and their Gaussian counterparts (g-bits) provide exactly this capability. 
These p-bits are binary stochastic units that naturally sample from Ising-type energy landscapes, and networks of p-bits implement Gibbs sampling of distributions defined by programmable couplings and biases. 
They thus constitute a form of \textit{Ising machine}~\cite{mohseni2022isingqubo, mcmahon2016coherentising}: hardware designed to minimize Ising Hamiltonians, but extended here to support controllable stochastic sampling rather than energy minimization alone.
Probabilistic computers built from p-bits have demonstrated strong performance in combinatorial optimization and machine learning~\cite{borders2019integer, aadit2022massively, niazi2024dbm}, with native hardware implementations on FPGAs and projected stochastic magnetic tunnel junction (sMTJ) devices offering orders-of-magnitude advantages in sampling throughput and energy efficiency over conventional GPU-based implementations~\cite{aadit2022massively} (quantified in Section~\ref{sec:pc_primitives}, Table~\ref{tab:platform_efficiency_comparison}).

The mapping of diffusion onto p-computers is particularly relevant because recent proposals for energy-based candidate selection~\cite{xu2025energybased} and Monte Carlo tree search~\cite{yoon2025montecarlotreediffusion, yoon2025fastmontecarlotree} during inference further increase the demand for fast, efficient probabilistic sampling. 

For physical systems governed by known interactions, such as Ising models with coupling coefficients $J_{ij}$, there is a further opportunity: the sampling distributions within diffusion can be constructed directly from the correlations of the underlying system. 
Rather than sampling from independent noise at each diffusion step, the stochastic component can perform structure-aware Gibbs dynamics that respect the physical couplings. 
We show that allowing diffusion to sample from these physically motivated distributions improves the accuracy of generated samples, and that independent diffusion is recovered as a special case when the couplings are set to zero ($J_{ij} = 0$).

\begin{figure*}[!t]
    \centering
    \includegraphics[width=0.6\textwidth]{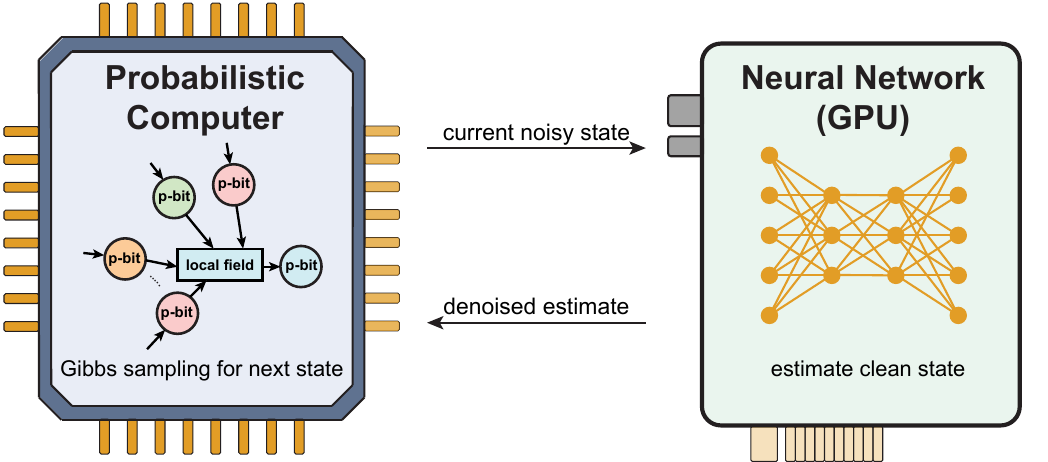}
    \caption{\textbf{Hybrid p-computer/GPU inference loop for correlated diffusion.}
    System-level view of one reverse-diffusion iteration. The current noisy configuration ($s_t$) is passed from the probabilistic computer to the neural network on the GPU, which produces per-site denoising probabilities $p=f_\theta(s_t)$. 
    A binary clean-state estimate $\hat{s}_0$ is then sampled from these probabilities and sent back to the probabilistic computer, where it initializes an ensemble of forward Gibbs chains under the known couplings $J_{ij}$ and the scheduled inverse temperatures. 
    Each chain produces a candidate for $s_{t-1}$, and the candidates are reweighted by the one-step likelihood $P(s_t\mid s_{t-1})$ before one is sampled to continue the reverse trajectory. 
    Repeating this procedure produces successive reverse states $s_t \rightarrow s_{t-1} \rightarrow \cdots \rightarrow s_0$, yielding the final sample $s_0$. The two blocks represent a conceptual division of labor between estimation and sampling, in an integrated CMOS+sMTJ realization both can reside on the same chip.
}
        \label{fig:fig1}
\end{figure*}

For systems with known couplings, direct MCMC sampling is available but can require prohibitively long chains to decorrelate, particularly near phase transitions or in frustrated systems. Correlated diffusion offers an alternative route: once trained, the hybrid neural-network/Gibbs reverse process can generate decorrelated approximate samples intended to match the target distribution without requiring long equilibrium chains at inference time. In this sense, the framework is analogous to a Boltzmann generator~\cite{noe2019boltzmann} for discrete systems, where the neural network proposes global denoising moves and the p-bit sampler enforces local consistency. Learned proposals have also been used to augment MCMC directly~\cite{albergo2019flow,nicoli2020unbiased,bunaiyan2025isingformer}, where a generative model whose sampling probability is exactly computable supplies global moves and an acceptance step makes the resulting chain asymptotically exact. Autoregressive networks have been applied directly to classical spin systems in the same spirit~\cite{biazzo2023arnn,biazzo2024sparse,delbono2025nearest,bialas2026priors,bialas2026transformers}, including the two-dimensional Edwards-Anderson model, where the network represents the joint distribution over all spins and so also provides an exact sampling probability. Correlated diffusion does not provide such a probability and is used here as a standalone generator rather than as an MCMC proposal; it embeds the physical structure directly into the generative dynamics instead, so the network estimates only per-site marginals of the clean state.

\begin{figure*}[!htbp]
    \centering
    \includegraphics[width =0.9 \textwidth ]{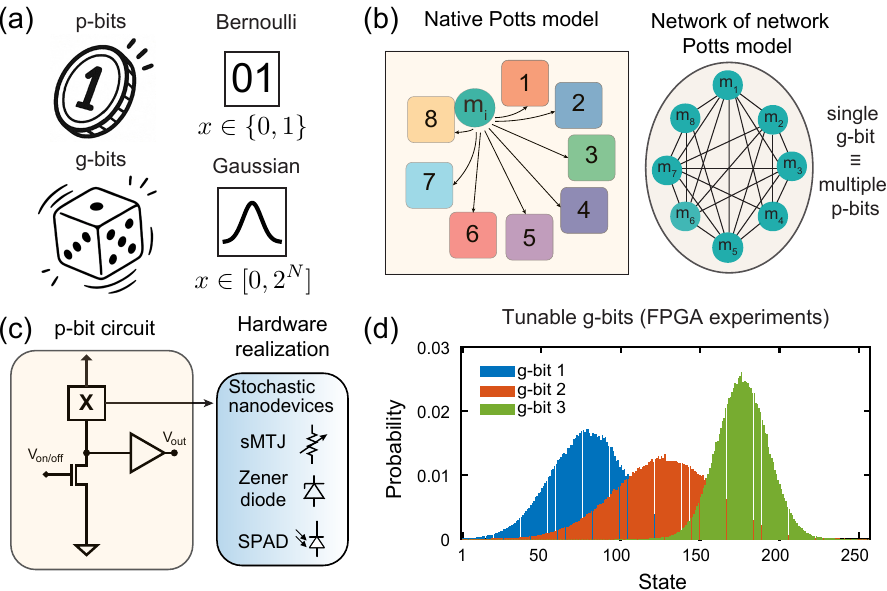}
    \caption{\textbf{Probabilistic computing hardware primitives.} (a) p-bits and g-bits serve as the building blocks for Bernoulli and Gaussian sampling, respectively. 
    (b) Multi-state Potts variables with integer states in $[0, 2^{N_p}-1]$ can be realized either as native Potts units or as networks of $N_p$ interconnected p-bits.
    (c) A p-bit circuit and candidate stochastic nanodevice realizations, including sMTJs, Zener diodes, and SPADs. 
    (d) Three tunable g-bits implemented on FPGA, each with 256 states and independently programmable means and variances.}
        \label{fig:fig2}
\end{figure*}

In this work, we formulate a generalized diffusion framework that incorporates Ising-structured Gibbs dynamics into both the forward noising and reverse inference processes.
In outline, the forward process starts from an equilibrium configuration and repeatedly applies a Gibbs sweep under the system's own couplings at a decreasing inverse temperature, progressively destroying structure until the state is disordered. Generation runs this backwards: at each reverse step a neural network proposes an estimate of the clean state from the current noisy configuration, short Gibbs chains started from that estimate produce candidates for the previous and less noisy configuration, and each candidate is reweighted by how likely it was to have produced the configuration actually observed.
The framework naturally maps onto a hybrid architecture (Fig.~\ref{fig:fig1}) in which p-computers handle the stochastic sampling and GPUs handle the neural-network-based conditional estimation.
This split is conceptual rather than physical: it shows a division of labor between conditional estimation and structured sampling, not necessarily two distinct devices. The p-bit sampling can be executed on the GPU itself~\cite{bunaiyan2025isingformer}, in which case the entire loop runs on a single node with no inter-device transfer, and our longer-term vision is an integrated CMOS+X substrate in which the sampler, for example sMTJs fabricated directly on CMOS~\cite{singh2024cmos}, and the neural-network accelerator share one chip. When a loosely coupled accelerator such as a PCIe-attached FPGA is used instead, the end-to-end benefit depends on whether the sampling speedup outweighs the communication overhead; the per-step interface traffic is in any case one configuration in each direction, since the candidate chains remain internal to the sampler.
More broadly, any Ising machine capable of controllable stochastic sampling (e.g., Ref.~\cite{mohseni2022isingqubo,bohm2022noise,ng2022efficient,suzuki2013chaotic,lee2025noise}) can serve as the sampling backend in this framework.
We introduce the probabilistic computing primitives, p-bits and g-bits (Fig.~\ref{fig:fig2}), that implement these dynamics, and quantify their sampling efficiency relative to GPUs (Table~\ref{tab:platform_efficiency_comparison}).
We demonstrate the approach on two benchmark systems: the 2D ferromagnetic Ising model and the 3D Edwards-Anderson spin glass, showing that correlated diffusion produces samples that align more closely with MCMC reference distributions of equilibrium observables.

\section{Probabilistic computing primitives: p-bits and g-bits}
\label{sec:pc_primitives}

Before developing the diffusion framework, we introduce the probabilistic computing primitives on which it is built.

The p-bit~\cite{camsari2019pbits, chowdhury2023fullstack} is a binary stochastic unit whose output $s_i\in\{-1,+1\}$ is sampled according to
\begin{equation}
\label{eq:pbit_update}
    s_i = \operatorname{sgn}\!\big[\tanh(\beta\, I_i) + \operatorname{rand}(-1,1)\big]
\end{equation}
where $\operatorname{rand}(-1,1)$ denotes a uniform random number on $(-1,1)$ and $\beta$ is the inverse temperature controlling the degree of stochasticity. Equivalently, the rule samples $s_i=+1$ with probability $\tfrac{1}{2}\big[1+\tanh(\beta I_i)\big] = \sigma(2\beta I_i)$, where $\sigma$ is the logistic function, so Eq.~\ref{eq:pbit_update} is an exact sampler for the Boltzmann conditional of spin $i$ given the current state of the others. The local field $I_i$ seen by each p-bit is determined by its neighbors through
\begin{equation}
\label{eq:pbit_field}
    I_i = \sum_j J_{ij}\, s_j + h_i
\end{equation}
where $J_{ij}$ are pairwise couplings and $h_i$ is an external bias. A network of p-bits connected by couplings $J_{ij}$ samples from the Ising energy
\begin{equation}
\label{eq:ising_energy}
    E(\mathbf{s}) = -\sum_{i<j} J_{ij}\, s_i\, s_j - \sum_i h_i\, s_i
\end{equation}
At high inverse temperature ($\beta\to\infty$), each p-bit becomes nearly deterministic, settling into the state favored by its local field; at low inverse temperature ($\beta\to 0$), the $\tanh$ term vanishes and the p-bit outputs become completely random. A decreasing $\beta$ schedule therefore provides a natural mechanism for progressively degrading a structured configuration into uncorrelated noise.

p-bits can be physically realized using a variety of stochastic nanodevices, including stochastic magnetic tunnel junctions (sMTJs)~\cite{borders2019integer,safranski2021demonstration,hayakawa2021nanosecond,schnitzspan2023nanosecond}, Zener diodes~\cite{patel2024pass}, and single-photon avalanche diodes (SPADs)~\cite{whitehead2024cmos} (Fig.~\ref{fig:fig2}c).

While p-bits are binary, many generative models, including Gaussian diffusion~\cite{dickstein2015diffusion} and Gaussian-Bernoulli Restricted Boltzmann Machines~\cite{kcho2011GBRBM,liao2022gaussianbernoullirbmstears}, require multi-state stochastic units. g-bits~\cite{singh2024gbits} fill this role: they are tunable Gaussian stochastic units that can be implemented either natively, using internal device physics to produce multi-state outputs directly, or as networks of interconnected p-bits that collectively represent $2^{N_p}$ states (Fig.~\ref{fig:fig2}), where $N_p$ denotes the number of p-bits internal to a single g-bit and is distinct from the total number of spins $N$ used throughout the rest of the paper. Networks of coupled g-bits sample from a quadratic energy landscape with programmable means, variances, and pairwise couplings, analogous to the Ising energy for p-bits (see Supplementary Section~I\,A for the g-bit energy and Gibbs sampling equations, and Supplementary Section~I\,B for the internal p-bit representation). Since diffusion models have two primary variants, Bernoulli and Gaussian~\cite{dickstein2015diffusion}, they map directly onto p-bits and g-bits, respectively, making p-computers broadly applicable to generative AI workloads.

\begin{table*}[htbp]
  \centering
  \caption{Comparison of efficiency for generating 32-bit random-number samples across hardware platforms}
  \label{tab:platform_efficiency_comparison}
  \resizebox{\textwidth}{!}{
    \begin{tabular}{ccccc}
      \toprule
      \textbf{Platform}
        & \textbf{Throughput (Gsample/s)}
        & \textbf{Power (W)}
        & \textbf{Efficiency (Gsamples/J)}
        & \textbf{Improvement factor} \\
      \midrule
      GPU (experimental)
        & \( 1000 \)
        & \( 280 \)
        & \( 3.5 \)
        & \( 1 \) \\
      FPGA (experimental)
        & \( 15000 \)
        & \( 25 \)
        & \( 600 \)
        & \( \sim 10^2 \) \\
      sMTJ (projected)
        & \( 31250 \)
        & \( 2 \)
        & \( 15625 \)
        & \( \sim 10^4 \) \\
      \bottomrule
    \end{tabular}
  }
\end{table*}

To quantify the hardware advantage, we compare the efficiency of 32-bit random sample generation across GPU, FPGA, and projected stochastic magnetic tunnel junction (sMTJ) platforms. We emphasize that these efficiency figures refer to raw stochastic sample generation, not full end-to-end diffusion inference, however, they are directly relevant for correlated diffusion because its reverse process shifts computational cost toward repeated Gibbs sampling. Throughout this section, a “sample” refers to one 32-bit random number produced by the underlying stochastic hardware primitive, not a full Ising configuration or a complete Gibbs sweep. The FPGA implementation uses 32-bit linear feedback shift registers (LFSRs) to generate uniformly distributed random numbers, an approach validated for MCMC sampling and optimization~\cite{aadit2022massively, singh2024cmos}.
GPU numbers are from our A100 experiments, informed by published benchmarks~\cite{spetko2021dgxa100,linford2022hpc,salmon2011parallel,Xu2015financerng}. FPGA numbers are from an AMD Alveo board, and sMTJ projections draw upon experimental results and scaling from Refs.~\cite{safranski2021demonstration,hayakawa2021nanosecond,schnitzspan2023nanosecond,borders2019integer}. All numbers are order-of-magnitude consistent across sources (see Supplementary Section~IV for details). In proposed projections, sMTJs can be fabricated directly on top of CMOS circuitry with minimal wiring delay, offering substantial area and energy advantages~\cite{singh2024cmos}.
As shown in Table~\ref{tab:platform_efficiency_comparison}, FPGA-based p-computers already achieve $\sim\!10^2 \times$ the sampling efficiency of GPUs, and projected sMTJ-based implementations extend this advantage to $\sim\!10^4\times$ over GPUs. Detailed hardware specifications are provided in Supplementary Section~III (Supplementary Table~S1), the FPGA p-bit architecture is described in Supplementary Section~II, and the GPU benchmarking methodology is detailed in Supplementary Section~IV. These efficiency gains become particularly significant for physics-based applications that require MCMC sampling. In the present implementation, one generated sample requires approximately
\[
N_{\mathrm{chains}}\sum_{t=1}^{T}(t-1)
\]
forward Gibbs sweeps. For $T=100$ and $N_{\mathrm{chains}}=10$, this corresponds to $49{,}500$ Gibbs sweeps per generated sample, making sampling with dedicated probabilistic computers directly relevant for reducing end-to-end inference costs. A step-by-step derivation of this sweep count is given in Supplementary Section~VII.

With these primitives in hand, diffusion steps can be interpreted as scheduled stochastic dynamics. A single p-bit with a time-dependent local bias $h_i^{(t)}$ and $J_{ij}=0$ implements a $2\times 2$ Markov transition corresponding to independent noise injection at each site. A network of $N$ interacting p-bits, by contrast, realizes a $2^N \times 2^N$ transition matrix that captures the full correlation structure of the Ising Hamiltonian.

In Section~\ref{sec:diffpbit}, we show how the independent diffusion emerges as the $J_{ij}=0$ limit, and in Section~\ref{sec:corrdiff}, we show how restoring the physical couplings yields a structure-aware correlated framework.

\begin{figure*}[!htbp]
    
    \includegraphics[width =1 \textwidth ]{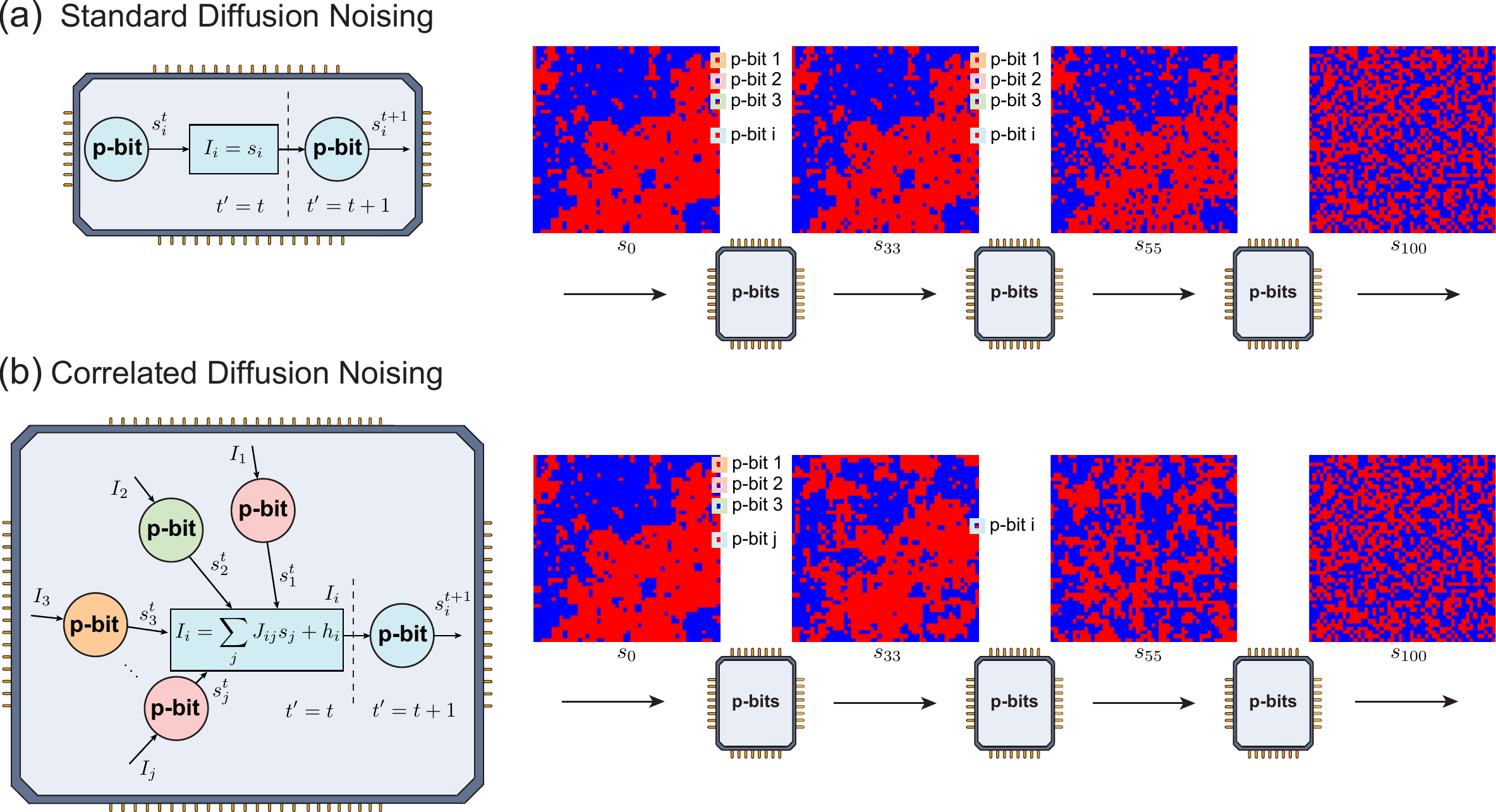}
    \caption{\textbf{Independent and correlated diffusion noising in a p-bit framework:}
    Illustration of the forward noising process implemented using probabilistic bits (p-bits), showing how structure-aware correlated noising generalizes the independent diffusion for a 2D ferromagnetic Ising system. (a) \emph{Independent diffusion noising}: each p-bit $s_i$ is updated with $J_{ij}=0$ under a time-dependent local bias $h_i=s_i^{t-1}$, so that $I_i=s_i^{t-1}$; a decreasing inverse temperature schedule $\beta\to 0$ (Eq.~\ref{eq:pbit_update}) progressively destroys correlations, resulting in factorized, site-wise noise injection (top row). (b) \emph{Correlated diffusion noising}: each p-bit is updated from an interaction-dependent local field $I_i=\sum_j J_{ij}s_j + h_i$, where explicit couplings $J_{ij}$ introduce spatially correlated noise consistent with the underlying Ising structure (bottom row). The independent noising process in (a) is recovered as a special case of (b) by setting $J_{ij}=0$ and absorbing the noising strength into the bias term $h_i$.}
    \label{fig:fig3}
\end{figure*}

\section{Diffusion as p-bit Dynamics}
\label{sec:diffpbit}

The forward process of diffusion where a structured state is progressively corrupted into noise is physically equivalent to heating a spin system toward infinite temperature. A network of p-bits with couplings $J_{ij}$ sampling 
from the Ising energy (Eq.~\ref{eq:ising_energy}) under a decreasing inverse temperature $\beta$ (Eq.~\ref{eq:pbit_update}) implements exactly this dynamics. We now show that the independent diffusion corresponds to the  $J_{ij}=0$ limit of this process, and that restoring couplings yields the correlated framework developed in Section~\ref{sec:corrdiff} (Fig.~\ref{fig:fig3}).

\paragraph*{Forward process.}
Consider first the independent case, $J_{ij}=0$. Each site carries a spin $s \in \{-1, +1\}$; at diffusion step $t$, the spin is flipped with probability $\eta_t$ and kept with probability $1 - \eta_t$. We use subscripts for the diffusion timestep on quantities that carry no site index ($s_t$, $\beta_{t-1}$, $\eta_t$) and superscripts when a site index occupies the subscript ($s_i^{t}$, $\beta_i^{t-1}$). This is the canonical forward noising model used throughout the discrete diffusion literature~\cite{dickstein2015diffusion}. Defining $a_t = 1 - 2\eta_t$ and the cumulative product $\lambda_t = \prod_{k=1}^{t} a_k$, the cumulative forward kernel from time $0$ to $t$ takes the compact form (see Supplementary Section~V\,A):
\begin{equation}
P(s_t \mid s_0) = \tfrac{1}{2}\big[1 + \lambda_t\, s_t\, s_0\big]
\end{equation}
Because each site evolves independently, the entire forward history from $0$ to $t$ compresses into the single scalar $\lambda_t$. This is the analytical convenience that independence buys: the $2^N \times 2^N$ transition matrix factorizes into $N$ identical $2\times 2$ kernels, each exponentiable in closed form.

\paragraph*{Reverse posterior.}
Given the cumulative kernel above, Bayes' rule yields a closed-form one-step reverse posterior (see Supplementary Section~V\,B):
\begin{equation}
P(s_{t-1} = +1 \mid s_t, s_0)
= \tfrac{1}{2} \left[ 1 +
\frac{a_t s_t + \lambda_{t-1} s_0}{1 + a_t \lambda_{t-1} s_t s_0} \right]
\end{equation}
Again, this closed form exists only because the forward chain is site-independent: $P(s_t \mid s_{t-1})$ and $P(s_{t-1} \mid s_0)$ are both scalar kernels, so the Bayesian inversion stays in $2\times 2$ space. In practice, $s_0$ is unknown and replaced by a neural-network prediction $\hat{s}_0 = f_\theta(s_t)$.

\paragraph*{Mapping to p-bit parameters.}
Diffusion models borrow the language of statistical mechanics, progressively noising and denoising configurations along a temperature-like schedule. For the independent case, this analogy can be made exact. The flip probability $\eta_{t-1}$ that defines the standard noising schedule can be recast as an inverse temperature for a p-bit network with $J_{ij}=0$. Each p-bit receives a time-dependent local bias $h_i^{(t)} = s_i^{t-1}$, so that $I_i^{(t)} = s_i^{t-1}$. Under inverse temperature $\beta_{t-1}$, the p-bit flip probability is
\begin{equation}
\Pr(s_i^t = -s_i^{t-1} \mid s_i^{t-1}) = \frac{1 - \tanh(\beta_{t-1})}{2}.
\end{equation}
Equating this to $\eta_{t-1}$ and inverting gives
\begin{equation}
\label{eq:etabetamap}
\beta_{t-1}=\arctanh(1-2\eta_{t-1}).
\end{equation}
As $\eta_{t-1}$ increases from $0$ to $0.5$, the inverse temperature drops from $+\infty$ to $0$: more noise corresponds to higher temperature, exactly as expected physically. The full derivation is in Supplementary Section~VI. This is the central equivalence of the section: independent diffusion is p-bit dynamics with $J_{ij}=0$ and a time-dependent local bias, where the entire noising schedule reduces to a temperature sweep.

\section{Structure-Aware Correlated Diffusion}
\label{sec:corrdiff}
The equivalence established in the previous section invites an obvious generalization: if independent diffusion corresponds to $J_{ij}=0$, what happens when the physical couplings are restored? For systems whose interaction structure is known, the factorized noise injection can be replaced with correlated MCMC dynamics that respect the underlying $J_{ij}$. 

Consider the simplest interacting case: two spins coupled by $J$.
The correlated forward process can be written exactly as a $4\times 4$ transition matrix $W_{\mathrm{Gibbs}}(\beta)$, constructed as the product of per-spin transition matrices in a fixed update order (see Supplementary Section~VII). Here the entry $[W_{\mathrm{Gibbs}}(\beta)]_{s',s}$ is the probability that one sweep at inverse temperature $\beta$ carries configuration $s$ to configuration $s'$, so the distribution over configurations evolves as $\pi_t = W_{\mathrm{Gibbs}}(\beta_{t-1})\,\pi_{t-1}$.
Because the matrix is small, quantities such as the prior and likelihood appearing in the reverse posterior (shown below in Eq.~\ref{eq:reverse_posterior}) can be computed by direct matrix multiplication.
For large $N$, however, the full $2^N \times 2^N$ transition matrix is intractable.
The scalable realization presented below replaces these exact matrix operations with Gibbs sampling using p-bit networks, where each spin is updated sequentially under its local field without ever constructing the full transition matrix.

\paragraph*{Correlated forward process.}
Restoring the couplings $J_{ij}\neq 0$ is straightforward: each p-bit is updated under the full local field (Eq.~\ref{eq:pbit_field}) rather than self-feedback alone. At each diffusion step $t$, a sequential Gibbs sweep~\cite{koller2009pgm} updates every spin once in a fixed order $\pi$, with each $s_i$ drawn from the p-bit update rule (Eq.~\ref{eq:pbit_update}) at inverse temperature $\beta_{t-1}$, conditioned on the current state of all neighbors. The same fixed order $\pi$ is used throughout the forward process and in the likelihood evaluation. The decreasing $\beta$ schedule plays the same role as before, progressively degrading structure into noise, but now each step performs one Gibbs sweep over the interacting Ising Hamiltonian rather than a product of independent sites. The independent process is recovered by setting $J_{ij}=0$ and retaining only the time-dependent local bias $h_i=s_i^{t-1}$. Gibbs sampling assumes only that each conditional distribution $P(s_i \mid s_{j\neq i})$ can be sampled exactly, which for an Ising model is precisely the p-bit update rule of Eq.~\ref{eq:pbit_update}; a sweep at fixed $\beta$ leaves the corresponding Boltzmann distribution invariant, although a single sweep does not by itself carry an arbitrary state to that distribution.

\paragraph*{One Gibbs sweep per diffusion step.}
Each diffusion step applies exactly one sweep, and this is a deliberate design choice rather than an approximation to equilibrium sampling at $\beta_{t-1}$. The forward process is required only to define a corruption kernel that degrades structure in a controlled way and whose one-step transition probability is known: the reverse posterior (Eq.~\ref{eq:reverse_posterior}) uses $P(s_t \mid s_{t-1})$ as its likelihood factor, and for a single sweep in the fixed order $\pi$ this factor is an exactly computable product of p-bit conditional probabilities. Performing many sweeps per step would instead compose that transition repeatedly and leave the one-step likelihood intractable, removing the quantity the reverse step needs without changing the purpose the forward process serves. A consequence is that the intermediate marginals are not the Boltzmann distributions at the corresponding $\beta_{t-1}$: the forward trajectory is a nonequilibrium annealing path, consistent with the non-equilibrium origins of diffusion models, whose endpoint as $\beta \to 0$ is the fully disordered state. Equilibrium is required only of the data at $t=0$, which is drawn from the equilibrium ensemble of the target Hamiltonian; the intermediate states serve as noisy conditioning inputs for the estimator and as the target of the one-step reverse kernel, and neither role requires them to be equilibrium samples.

\paragraph*{Training the conditional estimator.}
A key design choice is that the neural network does not learn the full joint reverse distribution over all spins. Instead, it is trained to predict independent per-site probabilities: given a noisy configuration $s_t$, the network outputs a probability $p_i = f_\theta(s_t)_i$ for each site $i$, representing the likelihood that the clean spin $s_{0,i} = +1$. Training minimizes a binary cross-entropy (BCE) loss between these per-site predictions and the true clean states $s_0$, summed over all sites. Each $p_i$ therefore approximates the marginal $P(s_{0,i}=+1 \mid s_t)$ of the clean-state posterior, and no joint structure over sites is represented in the network output; the explicit form of the loss and how it is estimated are given in Supplementary Section~VIII\,A.
In this work, the conditional estimator $f_\theta$ is implemented as a two-hidden-layer fully connected network.
Notably, $f_\theta$ receives only the noisy configuration $s_t$ and is not conditioned on the diffusion timestep $t$: all temporal structure enters exclusively through the Gibbs dynamics of the reverse kernel. We omit explicit timestep conditioning in order to isolate the effect of the correlated sampling kernel and use a single shared denoiser across all timesteps. A fuller discussion of this design choice, including how the timestep dependence is carried by the reverse Gibbs kernel through the $\beta$ schedule, is given in Supplementary Section~X.  The choice of network architecture is orthogonal to the correlated diffusion framework; we use a simple MLP here and leave exploration of more expressive architectures to future work.
Training hyperparameters are reported in Supplementary Table~S2.

Correlations between sites are introduced not by the network but by the reverse-kernel approximation. Given $s_t$, the network outputs per-site probabilities $p_i$, from which a single binary clean-state estimate $\hat{s}_0$ is sampled. This sampled $\hat{s}_0$ initializes an ensemble of forward Gibbs chains under the known couplings $J_{ij}$ and the scheduled inverse temperatures. The resulting candidate endpoints are reweighted by the one-step likelihood $P(s_t \mid s_{t-1})$, and one candidate is sampled to continue the reverse trajectory.

\begin{figure}[t!]
    \centering
    \includegraphics[width=\columnwidth]{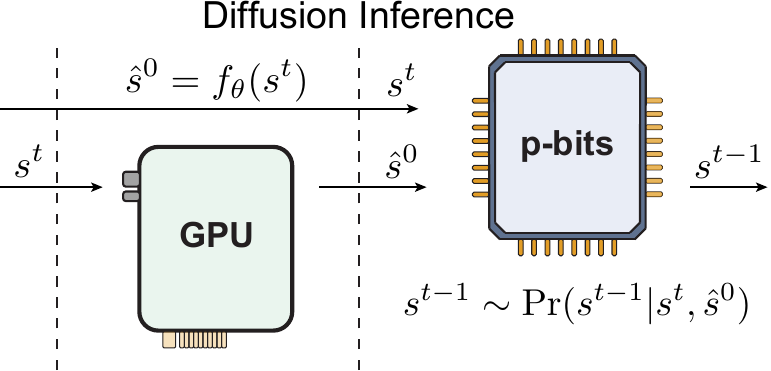}
    \caption{\textbf{Reverse generative process (diffusion inference) using neural-network guidance and p-bit sampling:}
Schematic of one reverse step. Given the current noisy state $s_t$, a neural network computes per-site denoising probabilities $p=f_\theta(s_t)$, from which a clean-state estimate $\hat{s}_0$ is sampled. An ensemble of forward Gibbs chains initialized at $\hat{s}_0$ then produces candidates for $s_{t-1}$ on the probabilistic computer. These candidates are reweighted according to
$\Pr(s_{t-1}\mid s_t,\hat{s}_0)\propto \Pr(s_t\mid s_{t-1})\,\Pr(s_{t-1}\mid \hat{s}_0)$,
and one candidate is sampled to continue the reverse trajectory. Repeating this procedure yields a sequence of progressively denoised states ending at $s_0$.}
    \label{fig:fig4}
\end{figure}

This division of labor is central to the framework (Fig.~\ref{fig:fig4}): the neural network provides a global denoising proposal, while the p-bit sampler restores the physically consistent correlation structure through Gibbs dynamics under the known interactions.

\begin{figure*}[htbp]
    \centering
    \includegraphics[width=0.8\textwidth]{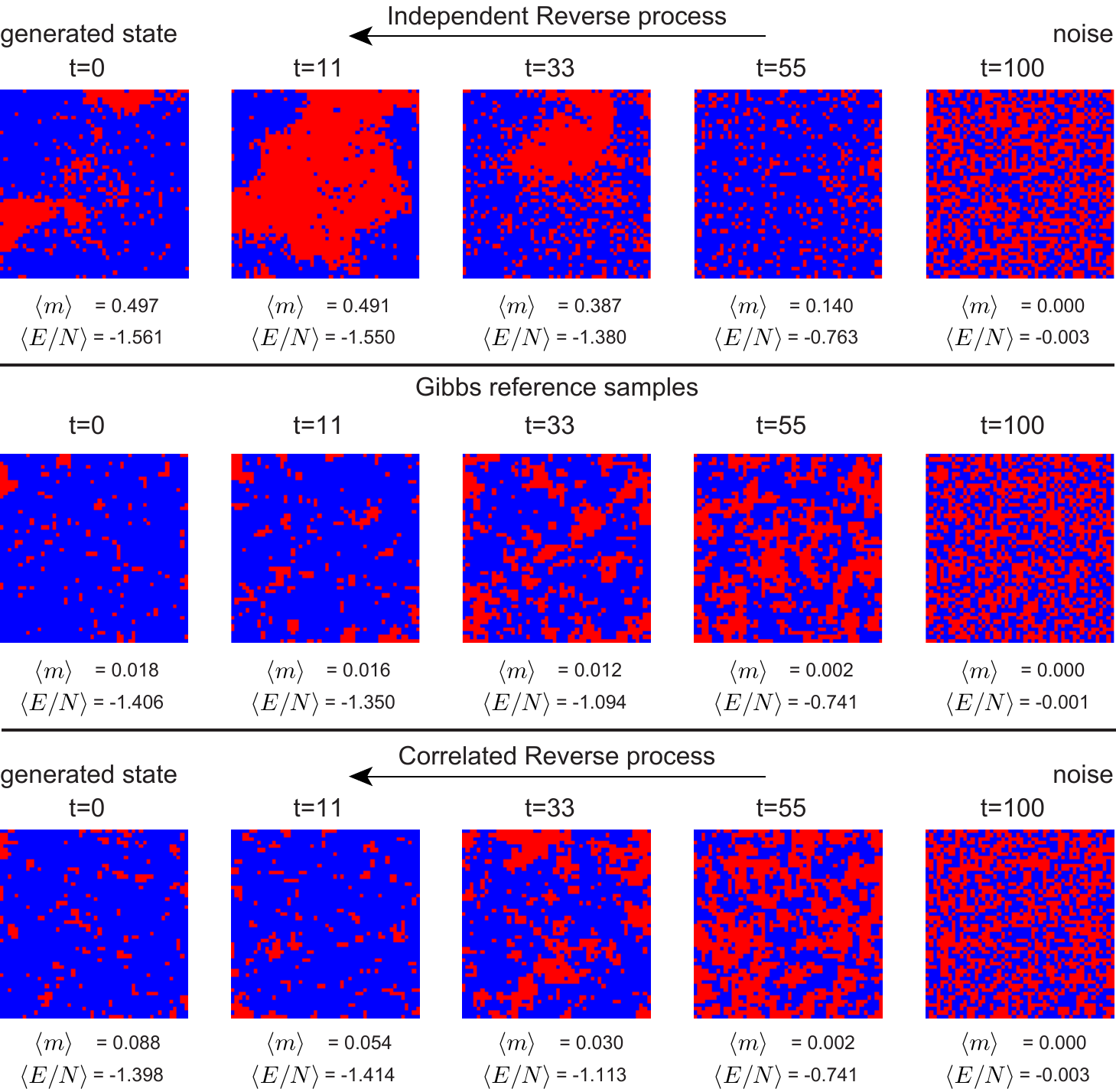}
    \caption{\textbf{Reverse trajectories for independent and correlated kernels on the 2D ferromagnetic Ising model.}
Representative configurations (right-to-left) at diffusion timesteps $t=\{100,55,33,11,0\}$ (red/blue denote $s=+1/-1$). \emph{Top:} independent (site-wise) kernel. \emph{Middle:} Gibbs-sampled reference at matching timesteps. \emph{Bottom:} correlated (interaction-aware) kernel. The mean magnetization $\langle m\rangle$ and mean energy per spin $\langle E\rangle/N$, averaged over the ensemble at each timestep, are reported beneath the corresponding column. The correlated kernel tracks the Gibbs reference more closely across timesteps, while the independent kernel produces spatially fragmented intermediate states and can deviate in magnetization even when energy appears comparable.}
    \label{fig:fig5}
\end{figure*}

\begin{figure*}[htbp]
    \centering
    \includegraphics[width =1 \textwidth ]{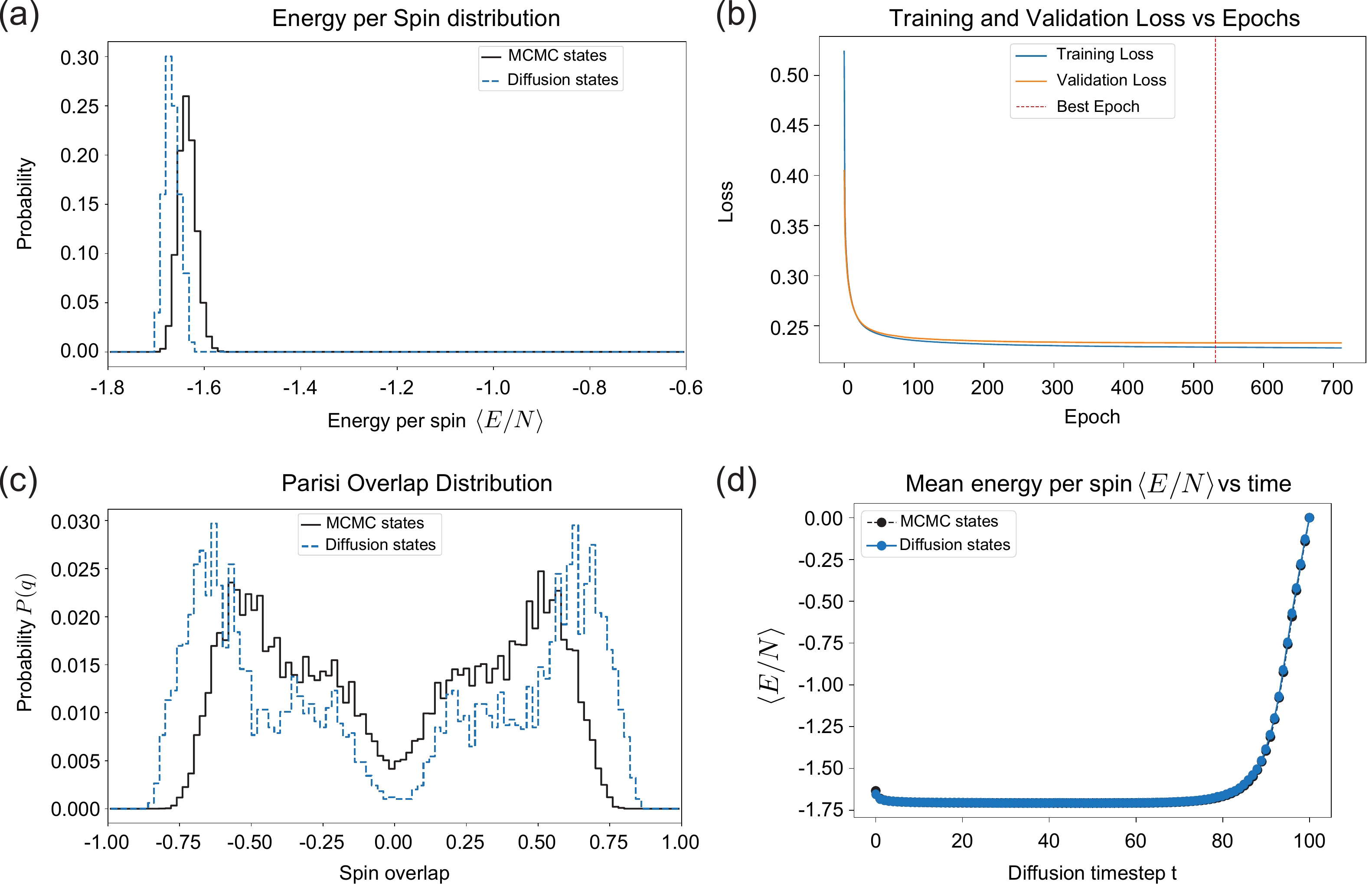}
   \caption{\textbf{Correlated diffusion applied to a 3D Edwards-Anderson spin glass near the phase transition.}
\textbf{(a)} Energy-per-spin distribution of the generated samples (blue) against the Monte Carlo reference (black). \textbf{(b)} Training and validation loss of the conditional estimator. \textbf{(c)} Parisi overlap distribution $P(q)$ of the generated samples (blue) compared with the equilibrium reference (black), showing that the framework captures the bimodal overlap structure of the spin-glass phase. \textbf{(d)} Mean energy per spin during the forward and reverse processes, for the generated trajectory (blue) and the Monte Carlo reference (black).}
    \label{fig:fig6}
\end{figure*}

\paragraph*{Correlated reverse process.}
The one-step reverse posterior has the same Bayesian structure as in the independent case, but with couplings restored, its evaluation becomes fundamentally different:
\begin{equation}
\label{eq:reverse_posterior}
P(s_{t-1} \mid s_t, s_0)
= \frac{P(s_t \mid s_{t-1})\; P(s_{t-1} \mid s_0)}{P(s_t \mid s_0)}
\end{equation}
The two factors in the numerator play distinct roles. The likelihood $P(s_t \mid s_{t-1})$ is the probability of reaching the current noisy state $s_t$ from a candidate $s_{t-1}$ via one forward Gibbs sweep in the fixed order $\pi$. Because the couplings $J_{ij}$ and inverse temperature $\beta_{t-1}$ are known, this term can be evaluated directly from the ordered one-sweep transition probabilities. The prior $P(s_{t-1} \mid s_0)$ represents the distribution over intermediate states $s_{t-1}$ obtained by running the forward chain from $s_0$ for $t-1$ steps. Unlike the likelihood, this term has no closed-form expression for interacting systems and must be estimated numerically via MCMC. During inference, $s_0$ is replaced by a sampled clean-state estimate $\hat{s}_0$ drawn from the neural-network output $f_\theta(s_t)$.

This marks the key computational departure from the independent case in Section~\ref{sec:diffpbit}: there, both the cumulative kernel $P(s_{t-1} \mid s_0)$ and the reverse posterior had closed forms because the $2\times 2$ kernels could be multiplied analytically. With $J_{ij}\neq 0$, neither is available, and both must be approximated via MCMC sampling.
Although the present reverse process is computationally intensive, this cost profile is precisely what makes the framework a natural match for probabilistic computers, whose native operation is fast Gibbs sampling.
Because dedicated probabilistic computers are being developed that can perform Gibbs sampling with 100s of billions of samples per second~\cite{aadit2022massively}, correlated diffusion is particularly well suited to such fast samplers.

In practice, the prior is approximated by running $N_{\mathrm{chains}}$ forward Gibbs chains from $\hat{s}_0$, and $s_{t-1}$ is selected by weighting each candidate by the one-step likelihood $P(s_t \mid s_{t-1})$.
Because the couplings $J_{ij}$ and inverse temperature $\beta_{t-1}$ are known, the one-step likelihood $P(s_t \mid s_{t-1})$ can be evaluated directly from the p-bit conditional probabilities (Eq.~\ref{eq:pbit_update}) along the Gibbs sweep. Drawing from the candidate ensemble with weights proportional to this likelihood realizes the posterior in Eq.~\ref{eq:reverse_posterior} without requiring explicit enumeration of all $2^N$ states.

\begin{algorithm}[t]
\caption{Approximate reverse step for correlated diffusion}
\label{alg:reverse_step}
\KwIn{$s_t$, $t$, $f_\theta$, $J_{ij}$, $\{\beta_\tau\}$, $\pi$, $N_{\mathrm{chains}}$}
Compute per-site probabilities $p = f_\theta(s_t)$\;
Sample $\hat{s}_0 \sim \prod_i \mathrm{Bernoulli}(p_i)$, mapped to $\{-1,+1\}$\;
\For{$c = 1$ \KwTo $N_{\mathrm{chains}}$}{
    Initialize $x_c^{(0)} \leftarrow \hat{s}_0$\;
    \For{$\tau = 1$ \KwTo $t-1$}{
        Gibbs sweep at $\beta_{\tau-1}$ with $J_{ij}$, order $\pi$ $\to$ $x_c^{(\tau)}$\;
    }
    $s_{t-1}^{(c)} \leftarrow x_c^{(t-1)}$\;
    $w_c \leftarrow P(s_t \mid s_{t-1}^{(c)})$\;
}
Aggregate duplicate candidates by summing weights\;
Sample $s_{t-1}$ from normalized weights\;
\end{algorithm}

\paragraph*{Reverse sampling protocol.}
Algorithm~\ref{alg:reverse_step} summarizes the concrete reverse-generation procedure used in all experiments, and each of its steps corresponds to a factor of Eq.~\ref{eq:reverse_posterior}. Lines 1 and 2 replace the unavailable clean state $s_0$ by a sampled estimate $\hat{s}_0$ drawn from the network output. The loop over $c$ generates the proposals: it draws $N_{\mathrm{chains}}$ independent samples from the prior $P(s_{t-1}\mid \hat{s}_0)$ by running the forward chain for $t-1$ sweeps from $\hat{s}_0$, this being the factor that has no closed form for interacting systems, and it is the only place where MCMC enters the reverse step. Each candidate is then assigned the weight $w_c = P(s_t \mid s_{t-1}^{(c)})$, the exactly computable one-sweep likelihood, which requires no sampling. Drawing a single candidate with probability proportional to $w_c$ is therefore self-normalized importance sampling from the posterior of Eq.~\ref{eq:reverse_posterior}, with the prior supplying the proposals and the likelihood supplying the weights. The aggregation step is exact bookkeeping rather than an approximation: identical candidate configurations represent one state whose weight is the sum of their individual weights. A step-by-step walkthrough is given in Supplementary Section~VII\,C.

\paragraph*{Scope and applicability.}
The framework as presented assumes a known coupling matrix $J_{ij}$, and this
assumption defines its scope. For a broad class of physics-based systems,
drawing equilibrium samples from an intractable Boltzmann distribution with a
\textit{given} Hamiltonian is itself the central computational problem,
including spin glasses, lattice models in statistical mechanics
where learned samplers are an active area of
research~\cite{noe2019boltzmann,albergo2019flow}, and
combinatorial-optimization problems whose Ising/QUBO encoding fixes $J_{ij}$ by
construction~\cite{mohseni2022isingqubo}. Correlated diffusion is designed for
this setting. Extending it to data whose interaction structure is unknown, such
as natural images, requires first inferring an approximate $J_{ij}$, for
example with a Boltzmann machine~\cite{niazi2024dbm} or inverse-Ising
estimation, which can then drive the correlated dynamics in place of a physical
Hamiltonian. We consider this a natural but untested extension and leave its
demonstration to future work. The advantage of structure-aware sampling is
largest when the couplings are known and the target is itself hard to sample
directly, as near criticality or under frustration, and when many independent
samples are required rather than a few.

\section{Results}
\label{sec:results}

Prior work has applied diffusion to Ising-type systems~\cite{lee2025thermogenising, bae2025dilutedising} and proposed iterative neural network inference as a means of capturing conditional structure in the reverse process~\cite{varma2025glaubergenerativemodeldiscrete}, but in all cases, correlations are captured implicitly by the network rather than explicitly encoded in the diffusion process, as proposed here.

\subsection{2D Ferromagnetic Ising Model}
\label{sec:results2d}

We first evaluate the proposed correlated diffusion framework on the 2D ferromagnetic Ising model on a $50\times 50$ lattice with open boundary conditions near criticality. 
The dataset consists of 10{,}000 equilibrium configurations collected via MCMC at zero external field and split 80/20 into training and validation sets.
Each $s_0$ is propagated through 100 forward diffusion transitions to produce noisy states $s_t$ at every timestep $t\in\{1,\ldots,T\}$, and a neural-network conditional estimator is trained on all resulting pairs $(s_t,s_0)$ to predict per-site denoising probabilities.
During inference, 100 reverse-generation trajectories are produced using $N_{\mathrm{chains}}=10$ candidate chains per reverse step, iterating the reverse kernel from a highly disordered $s_T$ back to $s_0$.

Fig.~\ref{fig:fig5} compares representative reverse trajectories for the independent and correlated kernels against Gibbs-sampled reference configurations at matching timesteps.
With a site-wise (independent) kernel, intermediate configurations exhibit spatially fragmented domains whose structure does not reflect the correlation length of the underlying Ising model at the corresponding temperature. In contrast, the interaction-aware (correlated) kernel produces a smoother evolution in which domains coarsen gradually as $t$ decreases, more closely resembling the Gibbs reference at each timestep.
The ensemble-averaged values of $\langle m \rangle$ and $\langle E/N\rangle$ reported beneath each column in Fig.~\ref{fig:fig5} make this contrast quantitative: the correlated kernel tracks the reference values across timesteps, while the independent kernel can deviate in magnetization even when energy appears comparable.
A more detailed analysis of energy and magnetization across the full reverse trajectory is provided in Supplementary Section~IX (Supplementary Fig.~S1): the correlated kernel tracks both $\langle E\rangle/N$ and $|\langle m\rangle|$ of the reference across all timesteps, whereas the independent kernel develops a systematic magnetization bias at small $t$ even where the energy appears comparable.
Finally, to compare the distributions of the final generated samples directly, we evaluate the two-point correlation function $C(r)$ of the generated $s_0$ ensembles for both kernels against the ideal reference on a smaller $5\times5$ system, where the correlated kernel reproduces the reference $C(r)$ closely and substantially better than the independent kernel (Supplementary Section~IX, Supplementary Fig.~S2).
Matching such ensemble observables at larger system sizes places correspondingly greater demands on the conditional estimator, and richer estimators, for example with explicit timestep conditioning, may improve the agreement further; we leave this to future work.

\subsection{3D Spin Glass}
\label{sec:results3d}

The 2D ferromagnet, with its uniform couplings and long-range order, provides a clean first test, but the real question is whether correlated diffusion can handle systems with frustrated, disordered interactions.
We therefore apply the framework to the 3D Edwards-Anderson spin glass on a $10\times10\times10$ cubic lattice ($N=1000$ spins) with 100 forward diffusion transitions. The reference and training configurations are equilibrium samples of this disorder instance at inverse temperature $\beta = 0.9091$ (temperature $1.10$), close to the critical temperature of the three-dimensional Edwards-Anderson model.
Here, random $\pm J$ couplings on a cubic lattice produce a rugged energy landscape with many competing low-energy states.
The dataset consists of 20{,}000 equilibrium configurations collected from a single disorder instance at this temperature, split 80/20 into training and validation sets. We report 100 reverse-generation trajectories using $N_{\mathrm{chains}}=10$ candidate chains per step.
The interaction matrix $J_{ij}$ is directly incorporated into the p-bit Gibbs dynamics during both the forward and reverse processes.

Fig.~\ref{fig:fig6} presents the results near the spin-glass phase transition.
The evolution of the mean energy per spin during the forward and reverse processes (Fig.~\ref{fig:fig6}d) shows that the reverse trajectory tracks the forward energy profile, confirming that the learned reverse process produces states at physically appropriate energy scales throughout the denoising sequence.
The energy-per-spin distribution of the generated samples (Fig.~\ref{fig:fig6}a) is close to the reference Monte Carlo data, with a visible offset in the peak position, indicating that the final generated configurations occupy the correct region of the energy landscape.

The most stringent test is the Parisi overlap distribution $P(q)$ (Fig.~\ref{fig:fig6}c), where $q = (1/N)\sum_i s_i^{(\alpha)} s_i^{(\beta)}$ measures the similarity between two independently generated configurations $\alpha$ and $\beta$.
Unlike energy, which is a single-configuration observable, $P(q)$ probes the statistical relationships across the generated ensemble: reproducing its characteristic spread near the phase transition requires that the model captures the coexistence of many equilibrium states with distinct overlap structure.
Reproducing the bimodal structure of the reference $P(q)$ therefore indicates that correlated diffusion captures the global structure of the spin-glass equilibrium, not just local energetics, although the generated distribution peaks at larger $|q|$ than the reference and is narrower than it.
A matched independent-diffusion baseline on the same instance, with the couplings removed from the diffusion kernels, is reported in Supplementary Section~XI (Supplementary Fig.~S3): the correlated kernel follows the reference $P(q)$ more closely, and in particular retains weight at intermediate overlap that the independent baseline does not.

Together, these results show that correlated diffusion with physically informed Gibbs dynamics generates samples that follow the equilibrium reference more closely than independent diffusion even in frustrated, high-dimensional systems where the energy landscape and the structure of the equilibrium state space are fundamentally more complex than in ordered magnets.

We briefly comment on the computational cost of correlated diffusion relative to direct MCMC:
Both methods use the same single-spin Gibbs kernel, so their costs compare directly in sweep counts. Correlated diffusion produces an independent sample
of the 3D spin glass in $N_{\mathrm{chains}}\,T(T-1)/2 = 49{,}500$ sweeps. Ensemble observables such as $P(q)$ additionally require many
independent chains. As in Boltzmann generators~\cite{noe2019boltzmann}, the
equilibrium sampling cost is paid once during training, after which
decorrelated samples are generated at fixed cost independent of the mixing
time; the approach pays off when many independent samples are required rather
than a few.

\section{Conclusion}
\label{sec:conclusion}
We have shown that independent diffusion is a special case of a more general framework in which the stochastic sampling component incorporates the interaction structure of the target system. The key observation is that independent diffusion corresponds to p-bit dynamics with $J_{ij}=0$: each site undergoes a temperature sweep under a time-dependent local bias, and the resulting $2\times 2$ kernels can be exponentiated analytically. Restoring the physical couplings $J_{ij}\neq 0$ replaces this factorized process with correlated Gibbs dynamics over the full Ising Hamiltonian, yielding a structure-aware diffusion framework in which the stochastic component respects known spatial correlations at every step.

On two benchmark systems, the 2D ferromagnetic Ising model and the 3D Edwards-Anderson spin glass, correlated diffusion produces samples that track MCMC reference distributions more closely than independent diffusion in both local observables (energy, magnetization) and global ensemble statistics (the Parisi overlap $P(q)$). These gains come from encoding physical knowledge directly into the diffusion dynamics rather than relying on the neural network to learn all correlations implicitly.

The framework naturally maps onto probabilistic computers built from p-bits, where the correlated Gibbs sweeps that independent diffusion avoids become the native operation. FPGA-based p-computers already provide $\sim\!10^2\times$ the sampling efficiency of GPUs, with projected sMTJ implementations reaching $\sim\!10^4\times$. As correlated diffusion shifts computational weight toward sampling, these hardware advantages become directly relevant.

Several directions remain open. The current reverse process relies on importance-weighted selection from a finite ensemble of Gibbs chains; more efficient sampling strategies, such as parallel tempering or learned proposal distributions, could reduce the number of chains required. The conditional estimator used here is a simple MLP: architectures that exploit spatial structure (e.g., graph neural networks) may further improve denoising accuracy. 

Finally, extending the framework beyond equilibrium Ising systems to non-equilibrium dynamics, continuous-state models via g-bits, or systems where the couplings $J_{ij}$ are partially known or must be inferred presents a natural next step. More broadly, the results suggest that the division of labor between neural-network estimation and structured stochastic sampling is a design axis that has been underexplored in the diffusion literature, and that dedicated sampling hardware can make previously impractical points along this axis accessible.

\section*{Acknowledgments}
N.S.S, M.M.R, S.N., and K.Y.C acknowledge support from the
National Science Foundation (NSF) CAREER Award under
grant number CCF 2106260, and the ONR-MURI grant
N000142312708. Use was made of computational facilities
purchased with funds from the National Science Foundation
(CNS-1725797) and administered by the Center for Scientific Computing (CSC). The CSC is supported by the California NanoSystems Institute and the Materials Research Science and Engineering Center (MRSEC; NSF DMR 2308708) at UC Santa Barbara.

\section*{Author contributions}
N.S.S. and K.Y.C. conceived the study. N.S.S. led the development of the correlated diffusion framework, performed the main numerical experiments, and wrote the initial version of the manuscript. M.M.R. contributed to the 2D ferromagnetic Ising benchmarks and to the correlated diffusion formulation in support of the main development. S.N. contributed to the hardware benchmarking analysis, including the FPGA-related evaluation. All authors discussed the results. N.S.S. and K.Y.C. finalized the manuscript.

\section*{Competing interests}
The authors declare no competing interests.

\section*{Data availability}
Equilibrium samples for the 2D ferromagnetic Ising model and the 3D Edwards--Anderson spin glass at criticality used in this study are available at \url{https://github.com/OPUSLab/CorrelatedDiffusion}.

\section*{Code availability}
Code for training and inference of both independent and correlated diffusion on the 2D ferromagnetic Ising model, and correlated diffusion on the 3D spin glass, including data loading, noising, training, inference, and metrics evaluation, is available at \url{https://github.com/OPUSLab/CorrelatedDiffusion}.

\bibliographystyle{unsrtnat}
\bibliography{references}

\balance


\onecolumngrid
\clearpage
\section*{Supplementary Information}
\beginsupplement

\section{g-bit formulation}\label{sec:diffalgo}

This section details the g-bit framework: Section~\ref{sec:intergbit} derives the Gibbs sampling equations for inter-g-bit interactions, and Section~\ref{sec:internalgbit} describes the internal p-bit representation of individual g-bits. Throughout Sections~I and~II we use $m_i \in \{-1,+1\}$ for p-bit states, consistent with the hardware literature; the diffusion sections use $s_i$ for spin states. The two are identical objects.

\subsection{Inter g-bit interactions}\label{sec:intergbit}

The p-bit energy is given by

\begin{equation}
\label{eq:GGBMLya}
E(m) = -\sum_{i<j} J_{ij}m_im_j - \sum_{i} h_i m_i
\end{equation}

similarly, when we work with g-bits, i.e., units that can represent multiple states that are gaussian-distributed, we have the energy given by

\begin{equation}
\label{eq:GGBMEnergy}
E(g) = \sum_{i} \frac{(g_i - b_i)^2}{2\sigma_{i}^2} - \sum_{i<j} \frac{W_{ij} g_i g_j}{\sigma_{i} \sigma_{j}}
\end{equation}

This section focuses on the Gibbs Sampling equations that approximate the energy terms corresponding to the Gaussian-Gaussian self and interaction terms.
Here we revisit our energy ansatz for the all g-bit network (all-Gaussian all-visible Boltzmann machine) which we will reference here as \ref{eq:GGBMEnergy}.

Here, we don't double count interactions, i.e., if the interaction between $g_1$ and $g_2$ is considered, we don't include the identical term for the $g_2$ and $g_1$ interaction.\\
\\
Next, consider a single-unit update in which unit $g_k$ changes its value from $g$ (current) to $g'$ (proposed), while all other units $g_{j \neq k}$ are held fixed. We write $\pi(g)$ and $\pi(g')$ for the probabilities of the two $N$-unit configurations $(g_1,\ldots,g_k{=}g,\ldots,g_N)$ and $(g_1,\ldots,g_k{=}g',\ldots,g_N)$, which differ only at unit $k$. We then apply the rule of detailed balance

\begin{equation}
\label{eq:ADB}
\frac{P(g \rightarrow g' \mid g_{j \neq k})}{P(g' \rightarrow g \mid g_{j \neq k})} = \frac{\pi(g')}{\pi(g)}\end{equation}

To devise a sampling algorithm, we are interested in $\displaystyle\frac{\pi(g')}{\pi(g)}$, which is given by Eq.~\ref{eq:A2},

\begin{equation}
\label{eq:A2}
\displaystyle\frac{e^{-E(g_k = g' \mid g_{j \neq k})}}{e^{-E(g_k = g \mid g_{j \neq k})}}
\end{equation}

Expression \ref{eq:A2} simplifies to the form $e^{-\Delta E}$, with $\Delta E = E(g_k{=}g' \mid g_{j \neq k}) - E(g_k{=}g \mid g_{j \neq k})$, where $-\Delta E$ is expressed as

\begin{equation}
\label{eq:A3}
-\Delta E = \left( \frac{(g - b_k)^2}{2\sigma_k^2} \right) -\left( \frac{(g' - b_k)^2}{2\sigma_k^2} \right) + \left( \sigma_k \sum_{j \neq k} \frac{W_{kj} g_j}{\sigma_j} \right) \left( \frac{g' - g}{\sigma_k^2} \right)
\end{equation}

We let $z_k = \sigma_k \sum_{j\neq k} \frac{W_{kj} g_j}{\sigma_j}$, which is not to be confused with the Boltzmann normalization factor, also known as the partition function. Further simplification steps are shown in \ref{eq:A4}, \ref{eq:A5}, \ref{eq:A6}, and \ref{eq:A7}.

\begin{equation}
\label{eq:A4}
-\Delta E = \frac{1}{2\sigma_k^2} \left[ (g - b_k)^2 - (g' - b_k)^2 + 2(g' - g) z_k \right]
\end{equation}

\begin{equation}
\label{eq:A5}
-\Delta E = \frac{1}{2\sigma_k^2} \left[ g^2 - g'^2 - 2 g b_k + 2 g' b_k + 2(g' - g) z_k \right]
\end{equation}

\begin{equation}
\label{eq:A6}
-\Delta E = \frac{1}{2\sigma_k^2} \left[\left(g^2 - 2 g (b_k + z_k) + (b_k + z_k)^2\right) - \left(g'^2 - 2 g' (b_k + z_k) + (b_k + z_k)^2\right)\right]
\end{equation}

\begin{equation}
\label{eq:A7}
\therefore \frac{P(g \rightarrow g' \mid g_{j \neq k})}{P(g' \rightarrow g \mid g_{j \neq k})} = \frac{\exp\!\left(\frac{-(g'-b_k-z_k)^2}{2\sigma_k^2}\right)}{\exp\!\left(\frac{-(g-b_k-z_k)^2}{2\sigma_k^2}\right)} = e^{-\Delta E}
\end{equation}

From Equation \ref{eq:A7}, we observe that the ratio is satisfied by a heat-bath (Gibbs) update in which the new value $g'$ is drawn, independently of the current value $g$, from the Gaussian density shown in Equation \ref{eq:A8}. We get a final expression for the mean and sigma as shown in Equation \ref{eq:A9}.

\begin{equation}
\label{eq:A8}
P(g \rightarrow g') = \frac{1}{\sqrt{2\pi\sigma^2}}\, e^{-\frac{(g' - \mu)^2}{2\sigma^2}}
\end{equation}

\begin{equation}
\label{eq:A9}
\begin{aligned}
   \mu &= b_k + \left( \sigma_k \sum_{j \neq k} \frac{W_{kj} g_j}{\sigma_j} \right), &
   \sigma &= \sigma_k
\end{aligned}
\end{equation}

Based on the expressions for the conditional probabilities, we frame a pair of Gibbs sampling equations. The first equation samples $g_i$
from the normal distribution with mean $\mu_i$ and standard deviation $\sigma_i$. The second equation determines the value of $\mu_i$ based on Equation \ref{eq:A9} to give the final iterative Gibbs sampling equations Equations \ref{eq:A10} and \ref{eq:A11}.

\begin{equation}
\label{eq:A10}
g_i \sim \mathcal{N} ( I_i, \sigma_i^2 )
\end{equation}

\begin{equation}
\label{eq:A11}
I_i = b_i + \sigma_i \sum_{j\neq i} \frac{W_{ij} g_j}{\sigma_j}
\end{equation}

Analogous to the inverse-temperature schedule used for p-bits, g-bit networks can be annealed by scaling the variance with $\beta$: $\sigma_i \leftarrow \sigma_i / \sqrt{\beta}$, so that decreasing $\beta$ broadens the sampling distribution toward the high-temperature (noisy) limit.

\subsection{G-bit internal representation}\label{sec:internalgbit}

A g-bit consists of a network of $N_p$ interconnected p-bits, where $N_p$ is the number of p-bits internal to a single g-bit (distinct from the number of spins $N$ in the main text). The energy of this Ising model is expressed as:
\begin{equation}
E = -E_0 \left( \frac{1}{2} \boldsymbol{m}^T J \boldsymbol{m} + \boldsymbol{h}^T \boldsymbol{m} \right)
\end{equation}
where $E_0$ is an overall energy scale that renders the couplings $J$ and biases $\boldsymbol{h}$ dimensionless.

Similar to the formulation described in Ref.~\cite{debashis2022gaussian}, we construct a random number ``$G$'' from the outputs of the p-bits, given by:

\begin{equation}
G = \boldsymbol{d}^T \boldsymbol{m}
\end{equation}

where $\boldsymbol{m}$ is the column vector of the $N_p$ p-bit states and

\begin{equation}
\boldsymbol{d} = \left(2^{a-1}, \, 2^{a-2}, \, \ldots, \, 2^0, \, 2^{-1}, \, 2^{-2}, \, \ldots, \, 2^{-b}\right)^T
    \label{eq:gend}
\end{equation}
where $a$ and $b$ denote the number of integer and fractional bits of the fixed-point representation, respectively, so that $N_p = a + b$.

We construct $X$ as a standardized (zero-mean, unit-variance) Gaussian variable from $G$, where $\mu$ and $\sigma$ denote the mean and standard deviation of the target Gaussian distribution to be programmed into the g-bit:

\begin{equation}
X = \frac{G - \mu}{\sigma}
\end{equation}

\begin{equation}
X = a' (\boldsymbol{d}^T \boldsymbol{m}) + b'
\end{equation}

where the constants $a'$ and $b'$ are given by

\begin{equation}
a' = \frac{1}{\sigma}
\end{equation}
\begin{equation}
b' = \frac{- \mu}{\sigma}
\end{equation}

The interaction matrix $J$ between the p-bits comprising the g-bit is set by $D = \boldsymbol{d}\boldsymbol{d}^{T} - \mathrm{diag}(\boldsymbol{d}\boldsymbol{d}^{T})$, the outer product of $\boldsymbol{d}$ with its diagonal set to zero. The derivation for $J$ and $\boldsymbol{h}$ follows from Ref.~\cite{debashis2022gaussian}, with $\boldsymbol{h}$ being the bias vector of the p-bit network.
\begin{equation}
J = -\frac{D}{\sigma^2}
\label{eq:Jgbit}
\end{equation}
\begin{equation}
\boldsymbol{h} = \frac{\mu \boldsymbol{d}}{\sigma^2}
\label{eq:hgbit}
\end{equation}

It is important to note that, unlike the continuous variable model which uses unipolar binary variables consistent with standard GBM architecture, the fixed-precision p-bit representation employs bipolar variables for all p-bits. Conversion between unipolar and bipolar variables can be easily made depending on the chosen mapping for the problem.

\section{FPGA Architecture}
\label{sec:FPGAarch}
     
The probabilistic computing architecture was implemented on an Xilinx Alveo U250 Data Center Accelerator Card using a custom digital design. 

Each p-bit is implemented using three main components: a pseudo-random number generator (PRNG), a nonlinear activation block, and a comparator. The PRNG produces uniformly distributed random numbers that introduce stochasticity into the p-bit updates. The activation function is implemented using a lookup table (LUT) that approximates the hyperbolic tangent function. The weighted input to each p-bit is computed using a multiply–accumulate (MAC) unit that sums the contributions from neighboring p-bit states and bias terms. The resulting activation value is compared with the random number to generate the updated binary p-bit state. The details of the architecture can be found in Ref ~\cite{niazi2024dbm}.

For efficient digital realization, the bipolar p-bit representation 
($m_i \in \{-1,1\}$) is mapped to a binary representation ($m_i \in \{0,1\}$). 
The coupling weights and bias terms are correspondingly transformed before being loaded onto the FPGA. To exploit hardware parallelism, p-bits are grouped using a graph-coloring scheme such that nodes within the same color can be updated simultaneously.
Communication between the host and FPGA is performed through a PCIe interface with a memory-mapped AXI register structure.

In this work, the FPGA p-computer is used to generate equilibrium training configurations via Gibbs sampling of the target Ising Hamiltonian.

\section{Hardware Specifications} 
\label{sec:hardbench}
\begin{table*}[htbp]
  \centering
  \caption{Hardware-specific parameters for GPU, FPGA, and sMTJ implementations. GPU literature values span several benchmarks~\cite{spetko2021dgxa100,linford2022hpc,salmon2011parallel,Xu2015financerng}; experimental values are from our A100 LFSR microbenchmark (Section~\ref{benchdet}).}
  \label{tab:all_device_specs}
  \makebox[\textwidth]{
    \begin{tabular}{clcc}
      \toprule
      \textbf{Platform} & \textbf{Specification}              & \textbf{Value}                  & \textbf{Unit} \\
      \midrule
      GPU (literature)  & GPU RNG throughput                 & \( 50 - 300 \)                  & Gsamples/s    \\
      GPU (literature)  & GPU power during RNG               & \( 200 - 400 \)                 & W             \\
      \midrule
      GPU (experimental)& GPU RNG throughput                 & \( 1000 \)                 & Gsamples/s    \\
      GPU (experimental)& GPU power during RNG               & \( 280 \)                  & W             \\
      \midrule
      FPGA              & Number of independent p-bits       & \( 10^{5} \)         &               \\
      FPGA              & Clock frequency                    & \( 150 \)         & MHz            \\
      FPGA              & Board power                        & \( 25 \)                        & W             \\
      \midrule
      sMTJ              & Number of sMTJs                    & \( 10^{6} \)         &               \\
      sMTJ              & sMTJ time constant (\( \tau \))    & \( 1 \)        & ns             \\
      sMTJ              & sMTJ frequency                     & \( 1 \)         & GHz            \\
      sMTJ              & Single sMTJ energy                 & \( 2 \)       & fJ             \\
      sMTJ              & Single sMTJ power                  & \( 2  \)        & $\mu$W             \\
      \bottomrule
    \end{tabular}
  }
\end{table*}

\section {Hardware Benchmarking details}\label{benchdet}

\subsection{GPU numbers}
The GPU numbers reported in Table~\ref{tab:platform_efficiency_comparison} (main text) are based on our own experiments on an NVIDIA A100-SXM4-80GB GPU, informed by several literature sources. Ref.~\cite{salmon2011parallel} describes how different GPU pseudo-random number generators (PRNGs) are emulated and implemented on hardware. Ref.~\cite{linford2022hpc} explicitly reports Gsamples/s throughput numbers for NVIDIA A100 GPUs. Diffusion workloads are typically compute-bound rather than memory-bound~\cite{chung2025mlenergybenchmark}, which allows power consumption to be estimated from compute-intensive benchmarks; Ref.~\cite{spetko2021dgxa100} provides power consumption data for A100 GPUs under such workloads. Together, these references form the basis for the literature-backed GPU estimate reported in Table~\ref{tab:all_device_specs} ($50$--$300$~Gsamples/s throughput, $200$--$400$~W power). Our experimental numbers are consistent in order of magnitude with these literature values and represent a best-case, fabric-local scenario on the A100. Both sets of numbers are listed in Table~\ref{tab:all_device_specs}.

Our experimental benchmark consists of a CUDA kernel in which each GPU thread maintains an independent 32-bit Galois linear feedback shift register (LFSR) whose state is stored entirely in registers. During kernel execution, each thread repeatedly advances its local LFSR and consumes the generated 32-bit samples within a tight loop, without writing any intermediate values to global or shared memory. This design intentionally eliminates memory traffic, ensuring that the measured throughput reflects local compute capability rather than memory bandwidth or host--device communication overhead.

A total of 552,960 concurrent threads are launched (2160 blocks with 256 threads per block), each generating 200,000 samples per iteration over 50 kernel iterations. Kernel execution time is measured using CUDA events, yielding an aggregate throughput of approximately $1.01 \times 10^{12}$ 32-bit samples per second. GPU power consumption is measured concurrently during the timed region by asynchronously sampling the on-board power sensors at 100 ms intervals, and the reported power corresponds to the time-averaged draw over the kernel execution window. The benchmark sustains an average power consumption of approximately 283~W, resulting in an energy efficiency of approximately 3.56~Gsamples/J. Since this represents a best-case fabric-local RNG scenario on a GPU, we use these experimental numbers as the GPU baseline in Table~\ref{tab:platform_efficiency_comparison}, providing the most favorable comparison point for the GPU and thus a conservative estimate of the FPGA and sMTJ improvement factors.

\subsection{FPGA numbers}

The FPGA numbers reported in Table~\ref{tab:platform_efficiency_comparison} are derived from the p-computer architecture described in Section~\ref{sec:FPGAarch}.
To isolate the random number generation capability, we strip away all logic, LUT, and MAC unit, and retain only the 32-bit LFSR-based pseudo-random number generators.
The architecture supports $10^5$ independent p-bits (Table~\ref{tab:all_device_specs}), each containing its own LFSR, clocked at 150~MHz.
Each LFSR produces one 32-bit sample per clock cycle, yielding an aggregate throughput of $10^5 \times 150 \times 10^6 = 1.5 \times 10^{13}$~samples/s, or approximately 15{,}000~Gsamples/s.
At a board power of 25~W, this gives an energy efficiency of 600~Gsamples/J, approximately $10^2\times$ the GPU baseline (Table~\ref{tab:platform_efficiency_comparison}).

\section{Independent Diffusion: Forward and Reverse Derivations}
\label{sec:indnoisedetails}

\subsection{Forward Process}
\label{sec:indnoisefwdsupp}

Each site carries a spin $s \in \{-1,+1\}$. At diffusion step $t$, the spin is flipped with probability $\eta_t$ and kept with probability $1-\eta_t$. Defining $a_t = 1-2\eta_t$ and the cumulative product $\lambda_t = \prod_{k=1}^{t} a_k$, the cumulative kernel from time $0$ to $t$ is

\begin{equation}
\begin{aligned}
P(s_t = s_0 \mid s_0) &= \tfrac{1}{2}(1 + \lambda_t), \\
P(s_t = -s_0 \mid s_0) &= \tfrac{1}{2}(1 - \lambda_t).
\end{aligned}
\end{equation}

As $t$ increases, $\lambda_t \to 0$ and both probabilities approach $\tfrac{1}{2}$, corresponding to complete randomization. The two cases above can be written compactly as
\begin{equation}
P(s_t \mid s_0) = \tfrac{1}{2}\big[1 + \lambda_t s_t s_0\big]
\end{equation}

\subsection{One-Step Reverse Posterior}
\label{sec:indnoiserevsupp}

The one-step reverse posterior follows from Bayes' rule:
\begin{equation}
P(s_{t-1} \mid s_t, s_0)
= \frac{P(s_t \mid s_{t-1}, s_0)\, P(s_{t-1} \mid s_0)}{P(s_t \mid s_0)}
\end{equation}

By the Markov property of the forward step, $P(s_t \mid s_{t-1}, s_0) = P(s_t \mid s_{t-1})$.
Using the one-step channel,
\begin{equation}
P(s_t \mid s_{t-1}) = \tfrac{1}{2}\big[1 + a_t s_t s_{t-1}\big]
\end{equation}

and the cumulative kernel up to $t-1$,
\begin{equation}
P(s_{t-1} \mid s_0) = \tfrac{1}{2}\big[1 + \lambda_{t-1} s_{t-1} s_0\big]
\end{equation}

we compute the probability for the event $s_{t-1} = +1$ (the other case is symmetric):
\begin{equation}
\label{eq:indrev_result_supp}
P(s_{t-1} = +1 \mid s_t, s_0)
= \tfrac{1}{2} \left[ 1 +
\frac{a_t s_t + \lambda_{t-1} s_0}{1 + a_t \lambda_{t-1} s_t s_0} \right]
\end{equation}

\section{Connecting Flip Probability (\texorpdfstring{$\eta$}{eta}) to Inverse Temperature (\texorpdfstring{$\beta$}{beta})}
\label{sec:etabetafull}

In independent diffusion models, the noising process involves updating a spin with spin-flip probability $\eta_{t-1}$ to transition from $s_i^{t-1}$ to $s_i^{t}$. This can be written as follows:

\begin{equation}
\begin{aligned}
&    \Pr\big(s_i^{t}=-s_i^{t-1}\mid s_i^{t-1})=\eta_{t-1}\\
&    \Pr\big(s_i^{t}=s_i^{t-1}\mid s_i^{t-1})=1-\eta_{t-1}
\end{aligned}
\end{equation}

Now, since this is an independent noising model, if we were to design a p-bit network that performs an analogous noising process, we enforce the following condition:

\begin{equation}
    I_{i,\text{ind}}^{t} = \beta_i^{t-1} h_i^{t-1} s_i^{t-1}
\end{equation}

Since we refer to an independent noising model, by definition, $I_{i,\text{ind}}$ cannot depend on other $s_j$. This essentially then forms a feedback, where the output of the p-bit is taken and fed back to the input after scaling by $h_i$.

Now, we try to determine how to compute $h_i^{t-1}$ given a flip probability $\eta_{t-1}$. We know that the p-bit flipping probability can be expressed in terms of our expression for $I_{i,\text{ind}}$ as follows, with $s\in\{-1,+1\}$:

\begin{equation}
    \Pr\big(s_i^{t}=s\mid s_i^{t-1}\big)=\frac{1+s\tanh(\beta_i^{t-1}h_i^{t-1} s_i^{t-1})}{2}
\end{equation}

Therefore,

\begin{equation}
    \Pr\big(s_i^{t}=+1\mid s_i^{t-1}=-1\big)=\frac{1-\tanh(\beta_i^{t-1}h_i^{t-1})}{2}
\end{equation}

Similarly,

\begin{equation}
    \Pr\big(s_i^{t}=-1\mid s_i^{t-1}=+1\big)=\frac{1-\tanh(\beta_i^{t-1}h_i^{t-1})}{2}
\end{equation}

Alternatively, we can write the flip event as ($s=-s_i^{t-1}$) follows:
\begin{equation}
\begin{aligned}
\Pr(s_i^t = -s_i^{t-1}\mid s_i^{t-1})
&= \frac{1 + (-s_i^{t-1})\tanh(\beta_i^{t-1} h_i^{t-1} s_i^{t-1})}{2} \\
&= \frac{1 - \tanh(\beta_i^{t-1} h_i^{t-1})}{2}.
\end{aligned}
\end{equation}

Now, we can equate this to $\eta$ and determine the expression for $h_i$ based on this.

\begin{equation}
    \eta_{t-1}=\frac{1-\tanh(\beta_i^{t-1}h_i^{t-1})}{2}
\end{equation}

Here, we can drop the subscript $i$

\begin{equation}
\beta_{t-1}h_{t-1}=\arctanh(1-2\eta_{t-1})
\end{equation}

Now, although $\beta_t$ and $h_t$ both appear, only one needs to be set based on $\eta$ and the other can remain fixed. Thus, we can frame this as a beta schedule entirely with a fixed bias ($h_t = 1$).\\

\begin{equation}
    \beta_{t-1}=\arctanh(1-2\eta_{t-1})
\end{equation}

Since in standard Bernoulli diffusion models $\eta_{t-1}$ (time-dependent, monotonically increasing) typically increases from $0$ (no noise) to $0.5$ (complete randomization), the corresponding inverse temperature ($\beta$) varies as:

\begin{equation}
    \beta_{t-1} = \operatorname{arctanh}(1 - 2\eta_{t-1}) \in [0, +\infty),
\end{equation}

indicating a progressively increasing noising process.

\section{Correlated Diffusion Model: Transition Matrix Approach}
\label{sec:corrnoisetrans}

This section presents the formulation for correlated diffusion using explicit transition matrices. While exact, this approach is limited to small systems due to the $2^N \times 2^N$ matrix size.

\subsection{Forward Process for Correlated Noising}
\label{sec:corrnoisetransfwd}

Let us consider two spins with a +$J$ coupling between them, and the energy of the pair given by \\
$E(s) = -  J s_1 s_2$ where $J > 0$.
\\

To perform noising or reverse annealing, during the forward process, as we move from $s_0$ to $s_t$, we decrease the value of beta to allow for a steady degradation of the state to noise.\\

A standard Gibbs sampling update for spin $i$ given the other spin $j$ is:

\begin{equation}
\Pr(s_i'=+1\mid s_j)=\frac{1+\tanh(\beta J s_j)}{2}=\frac{1+s_j\tanh (\beta J)}{2}.
\end{equation}
\\

Defining $q = \frac{1+\tanh(\beta J)}{2}$ and $p = 1 - q = \frac{1-\tanh(\beta J)}{2}$: if the neighbor is $+1$, $\Pr(s_i'=+1) = q$ and $\Pr(s_i'=-1) = p$; if the neighbor is $-1$, $\Pr(s_i'=+1) = p$ and $\Pr(s_i'=-1) = q$.\\

Now, we form a full step as first updating $s_1$ given $s_2$, then updating $s_2$ given the possibly changed $s_1$.

(For setting the matrix, let rows and columns go (-1,-1), (-1,+1), (+1,-1), (+1,+1), with MSB being $s_1$ and LSB being $s_2$.) \\

First, we update $s_1$ given $s_2$:

\begin{equation}
W(\beta)_{\text{1 $\leftarrow$ 2}}=
\begin{bmatrix}
q & 0 & q & 0\\
0 & p & 0 & p\\
p & 0 & p & 0\\
0 & q & 0 & q
\end{bmatrix}.
\end{equation}

Next, we update $s_2$ given $s_1$:

\begin{equation}
W(\beta)_{\text{2 $\leftarrow$ 1}}=
\begin{bmatrix}
q & q & 0 & 0\\
p & p & 0 & 0\\
0 & 0 & p & p\\
0 & 0 & q & q
\end{bmatrix}.
\end{equation}

Now, to evaluate the composite transition matrix across both update steps, we get $W(\beta)_{\text{Gibbs}}$ as follows.\\

\begin{equation}
W(\beta)_{\text{Gibbs}}=W(\beta)_{\text{2 $\leftarrow$ 1}} W(\beta)_{\text{1 $\leftarrow$ 2}}
\end{equation}

\begin{equation}
W(\beta)_{\text{Gibbs}}=
\begin{bmatrix}
q^2 & pq & q^2 & pq\\
pq & p^2 & pq & p^2\\
p^2 & pq & p^2 & pq\\
pq & q^2 & pq & q^2
\end{bmatrix}.
\end{equation}
\\

The forward process is defined using the diffusion-timestep-dependent $W_{\mathrm{Gibbs}}$ (note that timestep dependence is equivalent to $\beta$ dependence, since the $\beta$ schedule varies with time):

\begin{equation}
\begin{aligned}
& s_0 \text{ is the initial 2-spin state},\\
& \mathbf{v}_0 = \mathbf{onehot}(s_0) \\
& \text{For } t = 1,2,\ldots,T: \\
& \quad \pi_t = W_{\mathrm{Gibbs}}(\beta_{t-1}) \mathbf{v}_{t-1} \\
& \quad \mathbf{v}_t \sim \pi_t \\
\end{aligned}
\label{eq:corrfwdgen}
\end{equation}

\subsection{One-Step Reverse Posterior for Correlated Noising}
\label{sec:corrnoisetransrev}

We use the standard one-step Bayes formula to evaluate the posterior.

\begin{equation}
\Pr(s_{t-1}\mid s_t,s_0)=
\frac{\Pr(s_t\mid s_{t-1})\,\Pr(s_{t-1}\mid s_0)}{\Pr(s_t\mid s_0)}.
\end{equation}

\begin{equation}
\Pr(s_{t-1}\mid s_t,s_0) =
\frac{\ell_{s_t}\circ \pi_{t-1}}{Z}.
\end{equation}

Now, with the determined $W(\beta)_{\text{Gibbs}}$ we can look at how we compute the different terms in the Bayes formulation.

We can denote the \textbf{likelihood} by selecting the row from $W(\beta)_{\text{Gibbs}}$ corresponding to the $s_{t}$ value that we start the reverse process with. This gives us the likelihood of the observed $s_t$ from some $s_{t-1}$.

\begin{equation}
L_{s_t} = W(\beta_{t-1})_{\text{Gibbs}}[s_t, :]
\end{equation}

\begin{equation}
\Pr(s_t\mid s_{t-1}) = \ell_{s_t} = L_{s_t}^T
\end{equation}

The \textbf{prior} is $\pi_{t-1}$ obtained by pushing the given $s_0$ through the forward chain for $t-1$ steps. This forms our belief for $s_{t-1}$ only given $s_0$ and the forward noising schedule. $\mathbf{v}_0$ in this case is the one-hot encoded version of $s_0$ and has dimension of $4 \times 1$, and since our transition matrix is $4 \times 4$, the dimensions work out.

\begin{equation}
\begin{aligned}
\Pr(s_{t-1}\mid s_0) = \pi_{t-1}
&= W(\beta_{t-2})_{\text{Gibbs}}
   \times W(\beta_{t-3})_{\text{Gibbs}}
   \dots \\
&\quad \times W(\beta_1)_{\text{Gibbs}}
   \times W(\beta_0)_{\text{Gibbs}}\, \mathbf{v}_0
\end{aligned}
\end{equation}

The \textbf{normalization} is the total probability of seeing the current state $s_t$ under the forward model and our belief about $s_0$.

\begin{equation}
\begin{aligned}
\Pr(s_{t}\mid s_0)
&= [\,\pi_t\,]_{s_t} \\
&= [\,W(\beta_{t-1})_{\text{Gibbs}}
    \times W(\beta_{t-2})_{\text{Gibbs}}
    \dots \\
&\quad \times W(\beta_1)_{\text{Gibbs}}
    \times W(\beta_0)_{\text{Gibbs}}
    \,\mathbf{v}_0\,]_{s_t}
\end{aligned}
\end{equation}

The normalization factor can also be determined by the dot product of the previous two terms calculated, as shown below.

\begin{equation}
\text{Z} = L_{s_t}\cdot \pi_{t-1} = W(\beta_{t-1})_{\text{Gibbs}}[s_t, :] \cdot \pi_{t-1}
\end{equation}

So the final expression for the posterior can be given by the following:

\begin{equation}
\Pr(s_{t-1}\mid s_t,s_0)=
\frac{W(\beta_{t-1})_{\text{Gibbs}}[s_t, :]^T \circ \pi_{t-1}}{W(\beta_{t-1})_{\text{Gibbs}}[s_t, :] \cdot \pi_{t-1}}
\end{equation}

\subsection{Step-by-step reverse sampling}\label{sec:revsteps}

This section expands Algorithm~1 of the main text and identifies which factor of the reverse posterior each step estimates. The goal of one reverse step is to draw $s_{t-1}$ from a distribution proportional to $P(s_t\mid s_{t-1})\, P(s_{t-1}\mid s_0)$, in which the likelihood is available in closed form and the prior is not.

\textit{Step 1: clean-state estimate.} The true $s_0$ is unknown at inference time, so the network is evaluated once on the current state to give per-site probabilities $p_i = f_\theta(s_t)_i$, and a single binary configuration $\hat{s}_0$ is sampled from them, with $\hat{s}_{0,i}=+1$ with probability $p_i$. All subsequent steps condition on $\hat{s}_0$ in place of $s_0$.

\textit{Step 2: proposals from the prior.} The prior $P(s_{t-1}\mid \hat{s}_0)$ is the distribution of the forward process after $t-1$ steps started from $\hat{s}_0$. It has no closed form once $J_{ij}\neq 0$, so it is sampled: $N_{\mathrm{chains}}$ independent chains are each initialized at $\hat{s}_0$ and advanced by $t-1$ successive Gibbs sweeps at the scheduled inverse temperatures, in the same fixed order $\pi$ used in the forward process. The endpoint of chain $c$ is one candidate $s_{t-1}^{(c)}$, and these endpoints are samples from the prior by construction. This is the only place MCMC is used in the reverse step.

\textit{Step 3: likelihood weights.} Each candidate is weighted by $w_c = P(s_t \mid s_{t-1}^{(c)})$, the probability that one forward Gibbs sweep at $\beta_{t-1}$ carries $s_{t-1}^{(c)}$ to the observed $s_t$. Because that sweep updates every spin once in the fixed order $\pi$, this weight is the product of the corresponding p-bit conditional probabilities and is evaluated directly, with no sampling.

\textit{Step 4: selection.} Identical candidates are grouped and their weights summed, which is exact bookkeeping: a configuration produced by several chains is one state, and its weight is the total. One candidate is then drawn with probability $w_c/\sum_{c'} w_{c'}$ and returned as $s_{t-1}$.

Steps 2 to 4 together form a self-normalized importance sampler for the reverse posterior: proposals are drawn from the prior and reweighted by the likelihood, so the selected $s_{t-1}$ follows Eq.~\ref{eq:reverse_posterior} of the main text up to the Monte-Carlo error of a finite candidate ensemble, which is controlled by $N_{\mathrm{chains}}$. Repeating the four steps for $t=T,\ldots,1$ produces the full reverse trajectory ending at $s_0$.

\subsection{Gibbs Sweep Count for Correlated Reverse Sampling}
\label{sec:corrnoisesweepcost}

At each reverse step $t$, the state $s_{t-1}$ is drawn from the posterior
$\Pr(s_{t-1}\mid s_t,s_0)\propto \Pr(s_t\mid s_{t-1})\,\Pr(s_{t-1}\mid s_0)$.
The likelihood is read directly from a row of $W(\beta_{t-1})_{\text{Gibbs}}$ and
carries no sampling cost; the cost comes from the prior
$\pi_{t-1}=\Pr(s_{t-1}\mid s_0)$, which is the forward process advanced $t-1$
steps from $s_0$. For interacting systems this $2^N$ object has no closed form,
so it is estimated by Monte Carlo as in Algorithm 1 of the main text, drawing
$N_{\mathrm{chains}}$ forward Gibbs chains that each start from $\hat{s}_0$ and
apply $t-1$ sweeps. Since $t-1$ is the chain length and $N_{\mathrm{chains}}$ the
fixed number of chains, reverse step $t$ costs $N_{\mathrm{chains}}(t-1)$ sweeps
($t-1$ sweeps per chain, not $t-1$ chains), and the final reverse step $t=1$
costs none. Summing over the reverse trajectory,
\begin{equation}
\sum_{t=1}^{T} N_{\mathrm{chains}}(t-1) = N_{\mathrm{chains}}\,\frac{T(T-1)}{2},
\end{equation}
which for $T=100$ and $N_{\mathrm{chains}}=10$ gives $10 \times 4950 = 49{,}500$
forward Gibbs sweeps per generated sample, matching the main text.

\subsection{Sensitivity to the number of reverse chains}
\label{sec:nchains}

The candidate ensemble in Step 2 of Section~\ref{sec:revsteps} is finite, so the selection in Step 4 is a self-normalized importance sampler whose error decreases as $N_{\mathrm{chains}}$ grows. To test whether this finite ensemble accounts for the residual offset between the generated and reference distributions of the 3D Edwards-Anderson instance, we repeated inference with $N_{\mathrm{chains}}$ raised from $10$ to $50$, holding the trained estimator and all other settings fixed.

Neither the energy-per-spin distribution nor the overlap distribution $P(q)$ of the generated samples moved discernibly toward the reference. This suggests the finite candidate ensemble is not the dominant source of the residual discrepancy visible in Fig.~\ref{fig:fig6}, and we attribute the remaining error primarily to the clean-state estimator $f_\theta$.

\section{Network and Training Details}\label{sec:training_details}

\subsection{Training objective}\label{sec:trainobj}

The conditional estimator is trained by maximum likelihood on the per-site clean-state marginals. Writing $b_{0,i}=(s_{0,i}+1)/2\in\{0,1\}$ for the binary encoding of the clean spin at site $i$, and $f_\theta(s_t)_i\in(0,1)$ for the network output at that site, the loss is the binary cross-entropy
\begin{equation}
\label{eq:bceloss}
\mathcal{L}(\theta) = -\frac{1}{|\mathcal{D}|\,N}\sum_{(s_t,s_0)\in\mathcal{D}}\;\sum_{i=1}^{N}
\Big[\, b_{0,i}\,\log f_\theta(s_t)_i + (1-b_{0,i})\,\log\!\big(1-f_\theta(s_t)_i\big) \Big].
\end{equation}
The training set $\mathcal{D}$ consists of all pairs $(s_t,s_0)$ produced by the forward process, with $t=1,\ldots,T$, from every equilibrium configuration $s_0$ in the dataset, so each $s_0$ contributes $T$ pairs, one per diffusion timestep. The network input is the binary encoding $(s_t+1)/2$ of the noisy configuration. The expectation is thus approximated by an empirical average over this pooled set of pairs, and is estimated in practice by minibatch stochastic gradient descent with the Adam optimizer, drawing shuffled minibatches across all configurations and all timesteps (hyperparameters in Table~\ref{tab:training_details}). In the present implementation the pairs from all timesteps are pooled into this single average, so one estimator is shared across the schedule; conditioning the estimator on the timestep instead is a straightforward variant of the same objective.

\begin{table*}[b]
  \centering
  \vspace{-5pt}
  \caption{Architecture and training hyperparameters for the conditional estimator $f_\theta$, used identically for the 2D ferromagnetic Ising model and the 3D spin glass.}
  \label{tab:training_details}
  \begin{tabular}{lclc}
    \toprule
    \multicolumn{2}{c}{\textbf{Architecture}} & \multicolumn{2}{c}{\textbf{Training}} \\
    \cmidrule(lr){1-2} \cmidrule(lr){3-4}
    Input / output size & $N$ & Loss & Binary cross-entropy \\
    Hidden layers & 2 & Optimizer & Adam \\
    Hidden units per layer & 1024 & Learning rate & $10^{-6}$ \\
    Activation & ReLU & Weight decay & None \\
    Dropout & None & Batch size & 512 \\
    Layer normalization & None & Epochs & 4500 \\
     & & Early stopping patience & 180 \\
    \bottomrule
  \end{tabular}
\end{table*}

\textit{2D Ferromagnetic Ising Model:} Equilibrium configurations for the $50\times50$ ferromagnetic Ising model were generated via single-spin-flip Gibbs sampling.
For each independent trial the p-bit weights were first set to $\beta=0$, randomizing the initial state, after which the trial annealed through a fixed schedule of 81 equally spaced values running downward from $\beta=1.25$ to $\beta=0$, spending $10^5$ Gibbs sweeps at each value and reading out one configuration per $\beta$.
The 10{,}000 configurations used as the training dataset are the readouts at $\beta = 0.453125$, one from each of 10{,}000 independent trials.

\textit{3D Edwards-Anderson Spin Glass:} Reference configurations for the 3D Edwards-Anderson spin glass were generated via single-spin-flip Gibbs sampling.
All chains were initialized in the uniform $s_i = +1$ state.
Equilibration consisted of $10^6$ sweeps ramping linearly from $\beta = 0.1$ to $\beta = 0.9091$ (temperature $1.10$), close to the critical temperature of the three-dimensional Edwards-Anderson model, followed by $9\times10^6$ sweeps at that inverse temperature.
Each replica is an independent chain, and the 20{,}000 training configurations referenced in the main text are the final configurations of 20{,}000 such replicas after $10^7$ total sweeps.

\begin{figure*}[t]
    \centering
    \includegraphics[width=0.7125\textwidth]{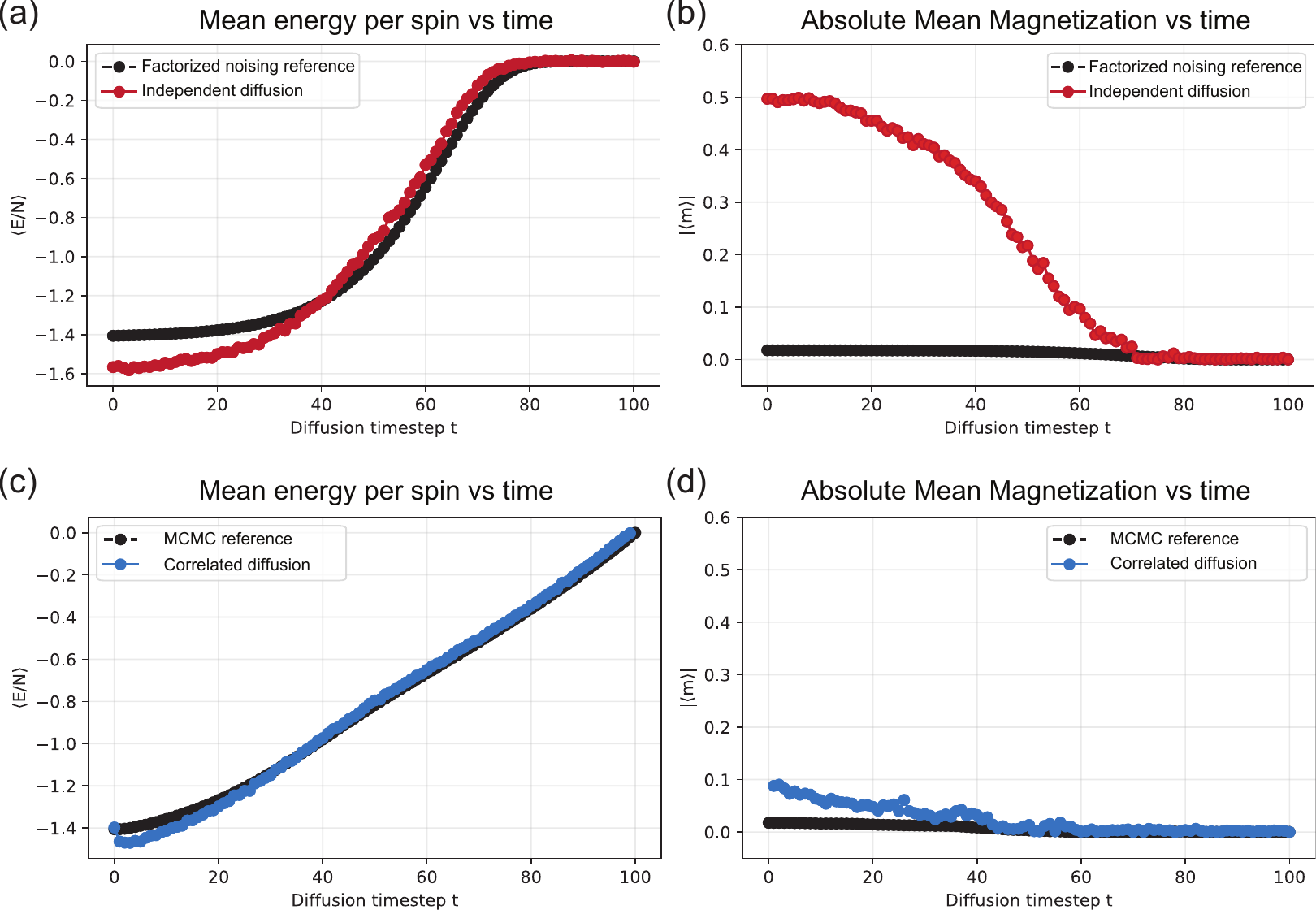}
    \vspace{-10pt}
   \caption{\textbf{Energy and magnetization across diffusion timesteps for generated Ising states.}
Mean energy per spin $\langle E\rangle/N$ and absolute mean magnetization $|\langle m\rangle|$ versus diffusion timestep $t$. \textbf{(a,b) Independent noising:} reverse-generated states (red) are compared with forward samples from the factorized noising kernel (black). Energy appears broadly similar under a chosen schedule (a), but magnetization differs strongly (b), showing that independent noising suppresses long-range order too rapidly. \textbf{(c,d) Correlated noising:} reverse-generated states (blue) are compared with interaction-respecting MCMC reference states (black). Both $\langle E\rangle/N$ and $|\langle m\rangle|$ track the reference more closely over $t$, consistent with the preservation of spatial correlations in the forward kernel.}
    \label{fig:figS1}\vspace{-10pt}
\end{figure*}

\section{2D Ferromagnetic Ising Model: Extended Results}\label{sec:isingresults}

Fig.~\ref{fig:figS1} extends the snapshot comparison in Fig.~\ref{fig:fig5} by plotting $\langle E\rangle/N$ and $|\langle m\rangle|$ across the full reverse trajectory.
Under independent noising, the energy trajectory roughly tracks the forward reference (Fig.~\ref{fig:figS1}a), but the magnetization develops a strong bias at small $t$ that is absent in the forward process (Fig.~\ref{fig:figS1}b).
Under correlated noising, both observables track the MCMC reference closely across all timesteps (Fig.~\ref{fig:figS1}c,d).

\begin{figure*}[t]
    \centering
    \includegraphics[width=0.7125\textwidth]{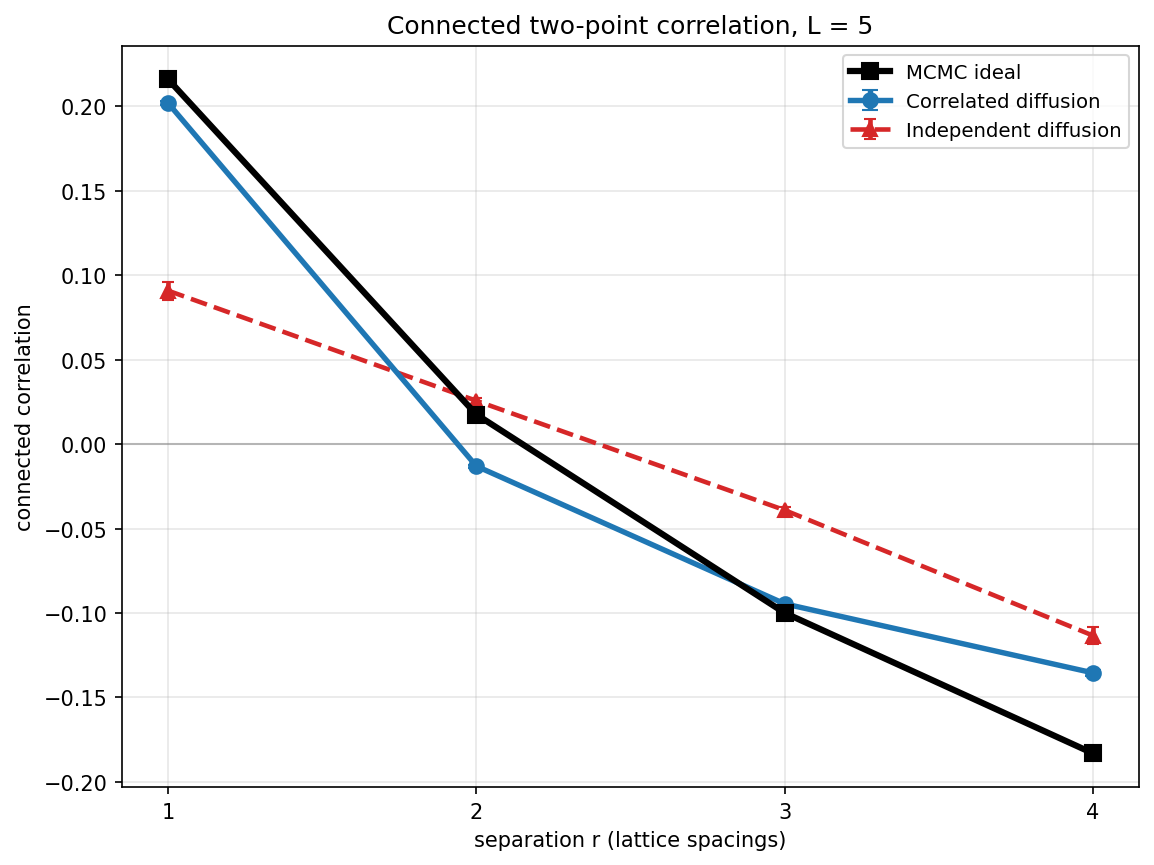}
    \vspace{-10pt}
   \caption{\textbf{Connected two-point correlation of the final generated samples ($L=5$).}
$C(r)=\langle s_i\, s_{i+r}\rangle-\langle m\rangle^{2}$ for the generated $s_0$ ensembles of the correlated (blue) and independent (red) kernels on a $5\times5$ ferromagnetic Ising model with open boundary conditions near the critical inverse temperature, compared against the ideal MCMC reference (black). Both kernels share the same network architecture and training protocol. The correlated kernel closely follows the reference, while the independent kernel deviates substantially. Error bars, where visible, denote the statistical uncertainty of the generated ensembles.}
    \label{fig:figS2}\vspace{-10pt}
\end{figure*}

To compare the distributions of the final generated samples directly, Fig.~\ref{fig:figS2} shows the connected two-point correlation $C(r)$ of the generated $s_0$ ensembles, averaged over all site pairs at separation $r$ along both lattice axes, for the correlated and independent kernels.
For this comparison we use a $5\times5$ lattice with open boundary conditions at the same inverse temperature.
Both kernels use the same network architecture, optimizer, training budget, and training data, so the comparison isolates the effect of the sampling kernel.
The correlated kernel reproduces the reference $C(r)$ closely (root-mean-square deviation $0.029$ over $r=1,\ldots,4$), while the independent kernel deviates substantially ($0.078$).
We use the smaller lattice for this test because matching ensemble observables of the generated distribution at $L=50$ places correspondingly greater demands on the conditional estimator (network capacity and conditioning); richer estimators, for example with explicit timestep conditioning (Section~\ref{sec:timestepfree}), may improve the agreement further and are beyond the scope of this work.

\section{On the Timestep-Free Conditional Estimator}
\label{sec:timestepfree}

The conditional estimator $f_\theta$ receives only the noisy configuration $s_t$
and is not conditioned on the diffusion timestep $t$. It is a clean-state
estimator, outputting per-site probabilities
$f_\theta(s_t)_i\approx P(s_{0,i}=+1\mid s_t)$, rather than a model of the full
reverse transition $P(s_{t-1}\mid s_t,t)$. The reverse update divides the work
between two components, with $f_\theta$ mapping $s_t$ to a clean-state estimate
$\hat{s}_0$ and the p-bit sampler mapping $(\hat{s}_0,s_t,J_{ij},\beta_{t-1})$ to
$s_{t-1}$ through the Bayesian reverse kernel of Eq.~\ref{eq:reverse_posterior}.
The timestep dependence is therefore not discarded but carried by the kernel: the
forward process applies $W_{\mathrm{Gibbs}}(\beta_{t-1})$ at each step
(Section~\ref{sec:corrnoisetrans}), and the reverse sampler is given the timestep
through $\beta_{t-1}$ even though $f_\theta$ is not.

In the independent-diffusion limit the cumulative kernel is
$P(s_t\mid s_0)=\tfrac{1}{2}[1+\lambda_t s_t s_0]$ with
$\lambda_t=\prod_k(1-2\eta_k)$, so the entire forward history compresses into the
single scalar $\lambda_t$ (Section~\ref{sec:indnoisedetails}) and conditioning on
it is directly useful, since the posterior depends on it explicitly. With the
couplings restored, the noising becomes a Gibbs sweep under the local field
$I_i=\sum_j J_{ij}s_j+h_i$ (independent diffusion is recovered at $J_{ij}=0$), and
the structured, timestep-dependent dynamics are supplied by the kernel rather than
the network. Because $f_\theta$ contributes only the clean-state estimate
$\hat{s}_0$, omitting the timestep is less consequential than it would be if the
network had to produce the full reverse transition.

We keep the estimator minimal so that the change we evaluate is in the sampler,
from independent to structure-aware p-bit sampling, rather than in the network.

The estimator $f_\theta$ could also embed timestep information directly, for
example through $\beta_t$ or $\lambda_t$ rather than the raw integer $t$. We note this as a natural extension for future work.

\section{3D Edwards-Anderson Spin Glass: Independent-Diffusion Baseline}\label{sec:eabaseline}

To isolate the contribution of the correlated kernel on the spin glass, we trained and ran an independent-diffusion baseline on the same Edwards-Anderson instance, in which the couplings are removed from the forward and reverse kernels ($J=0$) while every other component is held fixed: the same estimator architecture, loss, optimizer, training budget, reverse sampler, and the same $20{,}000$ equilibrium configurations.
The couplings are still used for all observables in both arms.
At $J=0$ the noise level is set by the per-spin flip probability $\eta$ rather than by $\beta$, and we use the same geometric $\eta$ schedule as the independent-limit case of the 2D ferromagnet.

\begin{figure*}[t!]
    \centering
    \includegraphics[width=0.7125\textwidth]{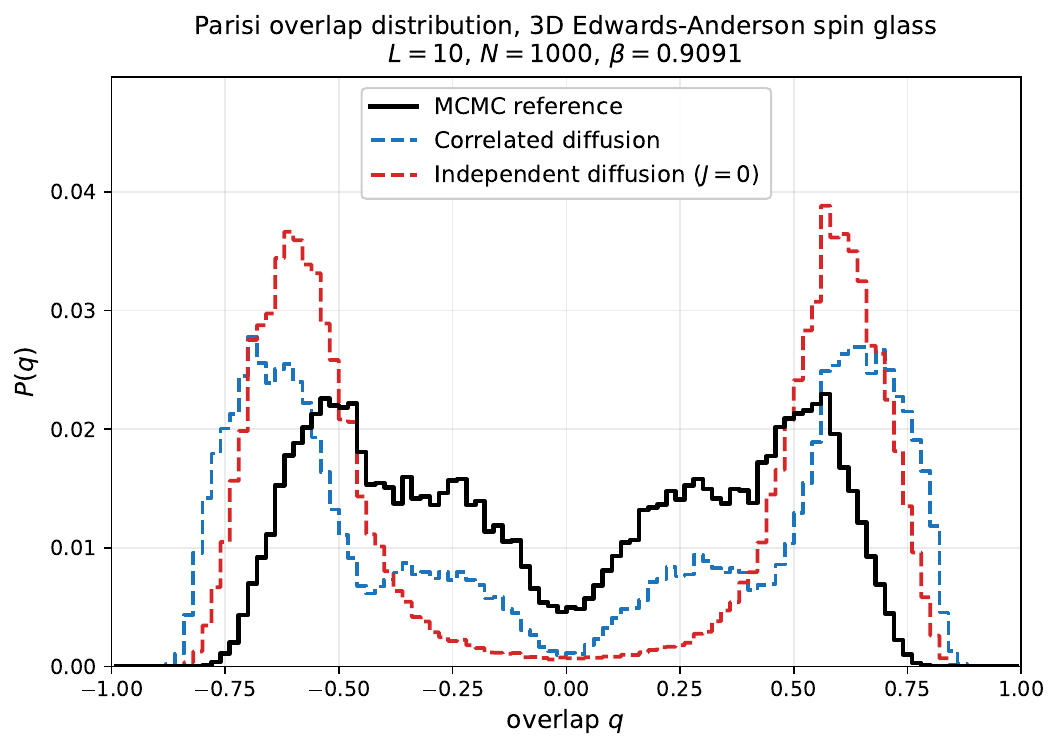}
    \vspace{-10pt}
   \caption{\textbf{Parisi overlap distribution on the 3D Edwards-Anderson instance: correlated versus independent diffusion.}
$P(q)$ of the generated ensembles for correlated diffusion (blue) and for an independent-diffusion baseline with the couplings removed from the diffusion kernels, $J=0$ (red), compared against the equilibrium MCMC reference (black), for the $10\times10\times10$ instance at $\beta=0.9091$. The two arms share the same estimator architecture, training protocol, reverse sampler, and training data, and differ only in whether the couplings enter the diffusion kernels. Neither reproduces the reference, but the correlated kernel retains weight at intermediate overlap that the independent baseline does not.}
    \label{fig:figS4}\vspace{-10pt}
\end{figure*}

Fig.~\ref{fig:figS4} compares the resulting overlap distributions.
Neither arm reproduces the equilibrium overlap distribution at this system size, but the correlated kernel follows the reference more closely than the independent baseline, in particular retaining weight at intermediate overlap where the independent baseline does not.

\section{Equilibration of the Reference Data}\label{sec:mixing}

Decorrelation was measured on 64 independent chains for the 2D system and 256 for the 3D instance.

\begin{figure*}[t!]
    \centering
    \includegraphics[width=\textwidth]{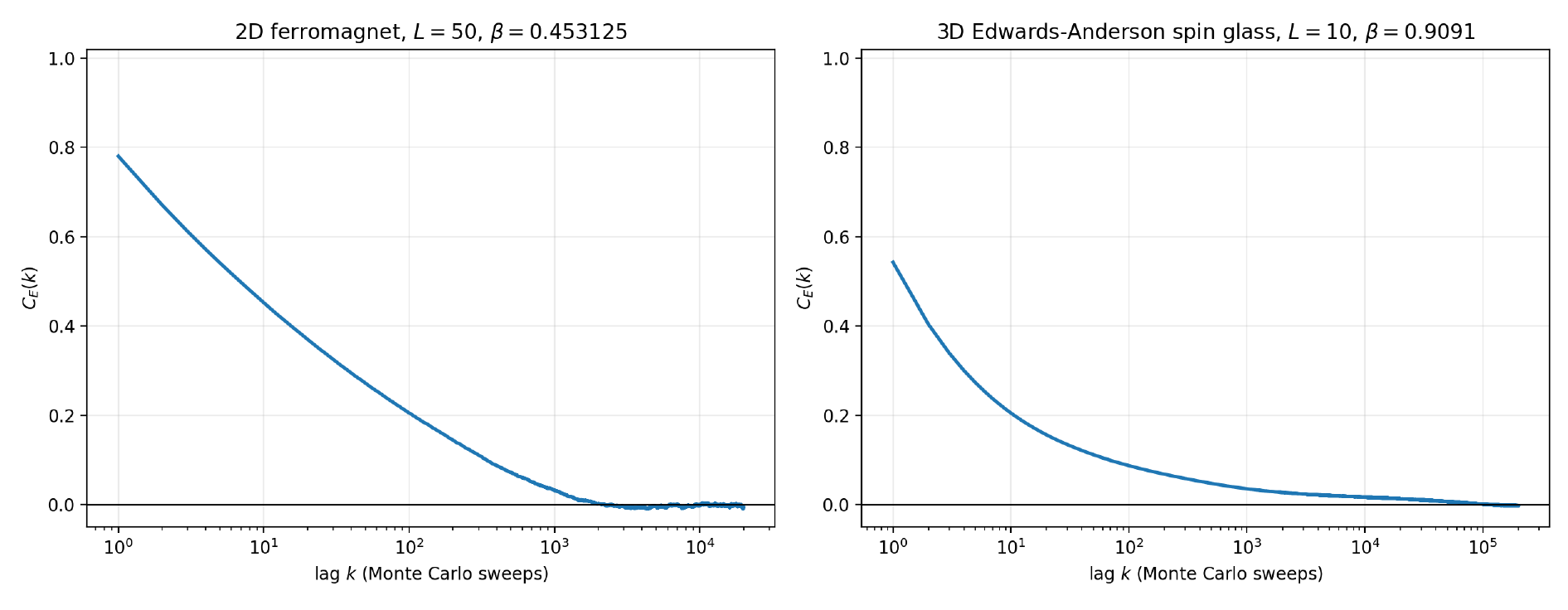}
    \vspace{-10pt}
   \caption{\textbf{Energy autocorrelation of the reference Monte Carlo dynamics.}
Normalized autocorrelation of the energy per spin, $C_E(k) = [\langle E(t)E(t+k)\rangle - \langle E\rangle^{2}]/[\langle E^{2}\rangle - \langle E\rangle^{2}]$, against lag $k$ in Monte Carlo sweeps, for the 2D ferromagnet at $\beta = 0.453125$ (left) and the 3D Edwards-Anderson instance at $\beta = 0.9091$ (right).}
    \label{fig:figS5}\vspace{-10pt}
\end{figure*}

Fig.~\ref{fig:figS5} shows the energy autocorrelation for both systems.
It decays to zero within about $2\times10^3$ sweeps in 2D and about $10^5$ sweeps in 3D.
The reference data were generated with $10^5$ sweeps at each $\beta$ in 2D and $10^7$ sweeps per replica in 3D, exceeding these lags by factors of about $50$ and $100$ respectively.
The delivered configurations come from independent chains rather than successive states of one chain, so correlation between samples is absent by construction and only within-chain equilibration is at issue.

\clearpage

\end{document}